\documentclass[11pt]{article}
\usepackage[final]{acl}
\usepackage{silence}
\usepackage{times}
\usepackage{latexsym}
\usepackage{amsmath}
\usepackage{amssymb}
\usepackage{amsfonts}
\usepackage[capitalise]{cleveref}
\crefname{figure}{Fig.}{Figs.}
\Crefname{figure}{Fig.}{Figs.}
\crefname{table}{Tab.}{Tabs.}
\crefname{section}{Sec.}{Sec.}
\Crefname{table}{Tab.}{Tabs.}
\usepackage{subcaption}
\usepackage{wrapfig}
\usepackage[T1]{fontenc}
\usepackage{enumitem}

\usepackage[utf8]{inputenc}

\usepackage{microtype}

\usepackage{inconsolata}

\usepackage{graphicx}
\usepackage{booktabs}
\usepackage{multirow}
\usepackage[skip=2pt]{caption}
\usepackage{alltt}

\usepackage[dvipsnames]{xcolor}
\usepackage{pifont}
\usepackage{tcolorbox}%
\usepackage{verbatim}%

\newcommand{\eg}{\textit{e.g.}}

\newcommand{\cmark}{\textcolor{OliveGreen}{\ding{51}}}
\newcommand{\xmark}{\textcolor{BrickRed}{\ding{55}}}
\definecolor{figgreen}{RGB}{0,128,0}
\definecolor{figteal}{RGB}{0,128,128}
\definecolor{figbrightred}{RGB}{192,0,0}
\definecolor{crimsonred}{RGB}{196,15,15}
\newcommand{\modelfullname}{Multi-Image MultImodal Retriever} 
\newcommand{\modelname}{MIMIR}
\newcommand{\datasetfullname}{Relational Entity Text-Image kNowledge Augmented}
\newcommand{\datasetname}{RETINA}
\newif\ifshowrevision
\showrevisionfalse
\ifshowrevision
  \newcommand{\added}[1]{{\color{red}#1}}
  \newcommand{\revcolor}{\color{red}}
\else
  \newcommand{\added}[1]{#1}
  \newcommand{\revcolor}{}
\fi

\title{Breaking the Visual Shortcuts in Multimodal Knowledge-Based Retrieval}

\author{
Dosung Lee\textsuperscript{1},
Sangwon Jung\textsuperscript{1,2},
Boyoung Kim\textsuperscript{1},
Minyoung Kim\textsuperscript{1}, \\
\textbf{Sungyeon Kim\textsuperscript{3}\thanks{This work was conducted independently and is not related to Amazon.},}
\textbf{Junyoung Sung\textsuperscript{1},
Wonpyo Park\textsuperscript{4},
Paul Hongsuck Seo\textsuperscript{1}} \\
\textsuperscript{1}Korea University, \textsuperscript{2}KAIST, \textsuperscript{3}Amazon, \textsuperscript{4}Google \\
{
\tt\small {\{dslee1219, bykimby, omniverse186, jys7451, phseo\}}@korea.ac.kr,
}
\\ 
{
\tt\small {s.jung@kaist.ac.kr}, 
\tt\small {sunykim@amazon.com},
\tt\small {wppark@google.com}
}
}

\begin{document}
\maketitle
\begin{abstract}
Existing Multimodal Knowledge-Based Visual Question Answering (MKB-VQA) benchmarks exhibit a \emph{visual shortcut}: the query image consistently depicts the main entity of the target document. 
We show that state-of-the-art multimodal retrievers perform comparably using the query image alone, revealing heavy reliance on this shortcut.
To address this, we introduce {\datasetname}, a new MKB-VQA benchmark free of visual shortcuts.
Unlike existing benchmarks, the visual query depicts a related entity of the target document rather than its main entity, with the textual query explicitly referencing it.
Since manual curation is difficult at scale, we propose a graph-guided pipeline that automatically constructs 120K training samples, complemented by 2K human-curated test samples.
Existing retrievers exhibit substantial performance drops on {\datasetname}, confirming their reliance on shortcut cues.
Motivated by this, we propose {\modelname}, which enriches document representations with images of multiple related entities.
Experiments show that {\modelname} mitigates the reliance on shortcut and outperforms on both {\datasetname} and prior benchmarks.
\end{abstract}

\section{Introduction}
\label{sec:intro}
Knowledge-Based Visual Question Answering (KB-VQA) requires reasoning beyond the image and its natural language query, drawing on external knowledge to answer correctly. 
Central to this task is a multimodal retriever that identifies the target document from the visual and textual query.

\begin{figure}[t]
    \centering
    \includegraphics[width=0.95\linewidth]{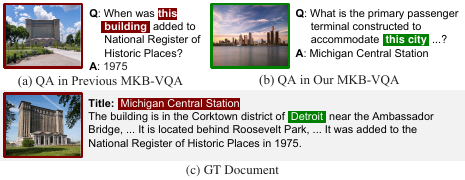}
    \caption{
    \textbf{Comparison of Previous MKB-VQA Benchmarks and {\datasetname}.} 
    (a) In previous benchmarks, the query image corresponds to the same visual entity (\textcolor{figbrightred}{\textbf{red}}) as (c) the shared GT document, creating a visual shortcut. (b) In {\datasetname}, the query uses a different visual entity (\textcolor{figgreen}{\textbf{green}}), breaking the visual shortcut.
    }
    \label{fig:task}
\end{figure}
\begin{figure}[t]
    \centering
    \begin{subfigure}{0.49\linewidth}
        \centering
        \includegraphics[width=0.9\linewidth]{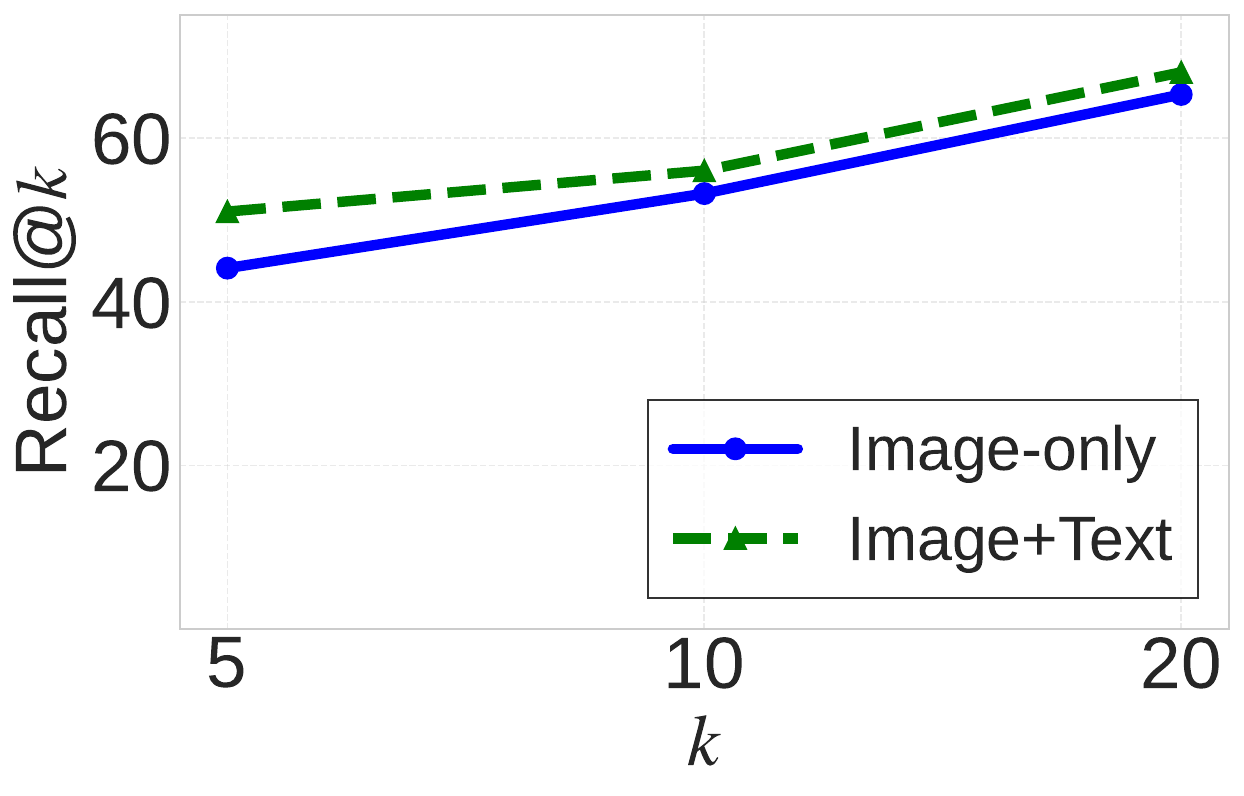}
        \caption{Infoseek}
        \label{fig:shortcut-infoseek}
    \end{subfigure}
    \hfill
    \begin{subfigure}{0.49\linewidth}
        \centering
        \includegraphics[width=0.9\linewidth]{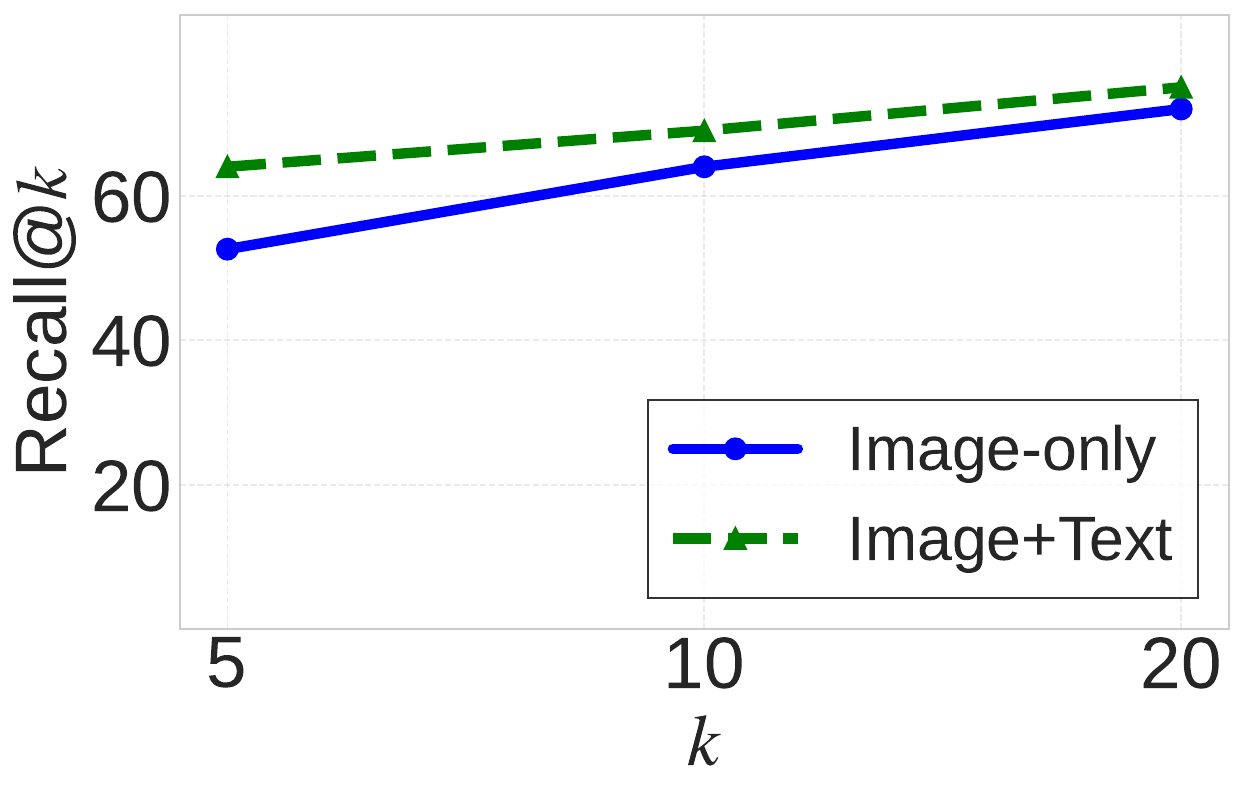}
        \caption{EVQA}

        \label{fig:shortcut-evqa}
    \end{subfigure}

    \caption{
        \textbf{Preliminary Experiment on Existing Benchmarks.}
        Recall@$k$ on benchmarks such as Infoseek~\cite{chen2023can} and EVQA~\cite{mensink2023encyclopedic},
        which exhibit visual shortcuts~\cite{deng-etal-2025-muka}.
        Models fine-tuned with image-only (\textbf{\textcolor{blue}{blue}}) queries versus image+text (\textbf{\textcolor{figgreen}{green}}) queries.
    }
    \label{fig:shortcut}
\end{figure}

However, existing MKB-VQA benchmarks~\cite{10.1145/3477495.3531753, chen2023can, mensink2023encyclopedic, shahMYP19} share a construction choice that induces a \textit{visual shortcut}: the query image depicts the same entity as the target document's main entity, so the target is largely retrievable through visual similarity alone, without utilizing the textual evidence.
As shown in \Cref{fig:task}a and \cref{fig:task}c, the query and document images depict the exact same entity—the building ``Michigan Central Station''—enabling trivial retrieval.
Our preliminary experiment confirms this shortcut exploitation: we trained a baseline retriever~\cite{deng-etal-2025-muka} without the textual query to isolate its reliance on visual cues.
As shown in \Cref{fig:shortcut}, the retriever achieves comparable performance using the query image alone. 
Moreover, in practice, relevant information often resides in related documents rather than those directly depicting the query image.
For instance, in \Cref{fig:task}b, given a photo of \textit{Detroit} and the question “Which cycling path traverses the dam of this city and extends through Cumbria county?'', the answer resides not in the “Detroit'' document but in the ``Michigan Central Station'' document—rendering visual cues alone insufficient for retrieval.

\begin{table*}[t]
    \centering
    \caption{\textbf{Comparison with Existing MKB-VQA Benchmarks.} {\datasetname} is the only benchmark in which the query image depicts a related entity rather than the main entity of the target document, eliminating the visual shortcut.}
    \setlength{\tabcolsep}{5pt}
    \scalebox{0.72}{
        \begin{tabular}{lcccccc}
        \toprule
        \textbf{Benchmark} & \textbf{\#Train} & \textbf{\#Test} & \textbf{Answer Type} & \textbf{Query Entity} & \textbf{Shortcut-Free} & \textbf{Auto Construction} \\
        \midrule
        KVQA~\cite{shahMYP19}                 & 165.7K & 18.3K & Entity      & GT Document Image & \xmark & \xmark \\
        ViQuAE~\cite{10.1145/3477495.3531753} & 1.2K   & 1.2K  & Span        & GT Document Image & \xmark & \xmark \\
        OVEN~\cite{hu2023opendomainvisualentityrecognition}                & 6M     & 24.9K & Entity      & GT Document Image & \xmark & \cmark \\
        InfoSeek~\cite{chen2023can}           & 676K   & 4.7K  & Entity/Span & GT Document Image & \xmark & \cmark \\
        EVQA~\cite{mensink2023encyclopedic}  & 167K   & 3.7K  & Span/Free   & GT Document Image & \xmark & \cmark \\
        \midrule
        \textbf{{\datasetname} (Ours)}        & 120K   & 2K    & Entity & \textbf{Related Entity Image} & \cmark & \cmark \\
        \bottomrule
        \end{tabular}}
    \label{tab:benchmark_comparison}
\end{table*}

To address this challenge, we construct {\datasetfullname} ({\datasetname}), an MKB-VQA benchmark with 120K training and 2K human-verified test samples, designed to eliminate visual shortcuts and better reflect real-world scenarios.
Prior benchmarks address a well-posed task in which the answer lies in the document of the depicted entity. 
{\datasetname} does not supersede them; it targets the complementary case, where the answer lies in a related document.
{\datasetname} serves as a counterbalance to existing shortcut-biased benchmarks.
Given the difficulty of manual construction, we build {\datasetname} via an automated LLM-driven pipeline. 
We introduce graph-based annotation as an intermediate step, leveraging graph structures to guide the LLM through relation extraction and query recomposition~\cite{ainslie2023gqa, tran2025reasonvqa, trivedi2022musiquemultihopquestionssinglehop, reddy-etal-2017-generating}. In particular, this allows us to decouple the complex generation process into structured sub-tasks, circumventing the reliance on direct prompting—a regime where LLMs are known to falter under compositional constraints~\cite{wen2024benchmarking} and suffer from degradation over long context windows~\cite{liu2023lostmiddlelanguagemodels}.
This strategy enables controlled generation by conditioning on target subgraphs of specific entities.

The same inductive bias is present on the model side, though it enters through the setup rather than the architectures.
Each document carries a single main-entity image, so that is what any retriever~\cite{deng-etal-2025-muka, lin2023finegrainedlateinteractionmultimodalretrieval, long2025retrieval, wei2024uniir, yu2025cafeunifyingrepresentationgeneration, lin2024preflmr} has to encode.
We confirm this inductive bias experimentally.
On {\datasetname}, existing retrievers tend to rank documents whose main-entity image matches the query above the related document that contains the answer.
We propose {\modelfullname} ({\modelname}), a retrieval framework that relaxes this inductive bias by representing each document with images of multiple related entities rather than a single main-entity image.
Specifically, {\modelname} extracts related entities from each document and collects their images, which are fused into the document embedding to guide retrieval when the query depicts one of these entities.
% [Original] This multi-image augmentation enables retrieval even without visual shortcuts.
\added{What matters is not the number of indexed images but their relation to the document: with the index size held fixed, neither additional views of the main entity nor images of unrelated entities recover the gain, which arises only when the indexed images depict the document's related entities (\cref{sec:capacity}).}
Experiments show that {\modelname} nearly doubles recall on {\datasetname} while matching or improving on existing MKB-VQA benchmarks.
Our contributions are as follows:
\begin{itemize}[itemsep=2pt, topsep=3pt, parsep=0pt, leftmargin=*]
\item We identify that visual shortcuts let retrievers recover the target from the query image alone in existing MKB-VQA benchmarks;
\item We propose an automatic graph-based pipeline that constructs shortcut-free samples;
\item We introduce {\datasetname}, a human-curated benchmark that counterbalances prior ones, enabling holistic evaluation with and without shortcuts;
\item 
% [Original] We propose {\modelname}, a retrieval framework that addresses the 
% [Original] visual shortcut.
\added{We propose {\modelname}, a retriever that identifies each document's related entities and indexes their images, and show that its gains stem from this entity association rather than from added index capacity.}
\end{itemize}
\added{The benchmark and the method form a single argument: {\datasetname} reveals the failure, and {\modelname} confirms its cause with a minimal fix, showing that knowledge about related entities must be made visually matchable at the document level.}

\section{Related Works}
\label{sec:related_works}
\begin{figure*}[t]
    \centering    
    \includegraphics[width=\linewidth]{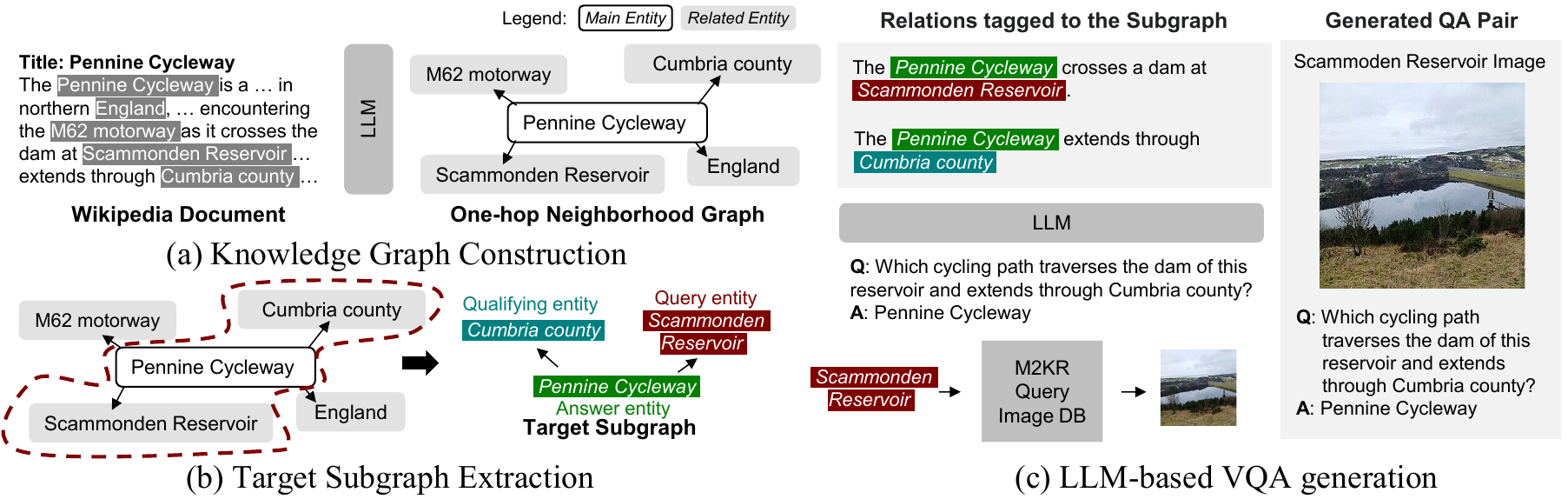}
    \caption{
    \textbf{{\datasetname} Generation Pipeline.} (a) constructs one-hop neighborhood graphs by extracting named entities (\textbf{\textcolor{gray}{gray}} box) related to the answer entity (white box) and their relations using an LLM (relations are omitted); (b) samples a query entity (\textcolor{figbrightred}{\textbf{red}}) and qualifying entity (\textcolor{figteal}{\textbf{teal}}) to form a target subgraph with the answer entity (\textcolor{figgreen}{\textbf{green}}); and (c) feeds the target subgraph into an LLM to generate a textual query and collect a corresponding image from M2KR Images~\cite{lin2024preflmr}. The query is then paraphrased to minimize lexical overlap with the document.}
    \label{fig:pipeline}
\end{figure*}
\noindent \textbf{Knowledge-Based VQA Benchmarks} 
Knowledge-based VQA datasets can be categorized by the modality of their external knowledge bases. Benchmarks relying solely on textual knowledge include SK-VQA~\cite{su2024sk}, S3VQA and OKVQAS3~\cite{jain2021select}, OK-VQA~\cite{marino2019okvqavisualquestionanswering}, A-OKVQA~\cite{schwenk2022aokvqabenchmarkvisualquestion}, and FVQA~\cite{wang2017fvqa}, requiring models to retrieve and reason over textual documents. A second group employs multimodal knowledge bases pairing documents with images, including KVQA~\cite{shahMYP19}, ViQuAE~\cite{10.1145/3477495.3531753}, Infoseek~\cite{chen2023can}, and E-VQA~\cite{mensink2023encyclopedic}; M-BEIR~\cite{wei2024uniir} also features a multimodal knowledge base but targets instructed retrieval rather than KB-VQA. Finally, some datasets integrate diverse modalities: M2KR~\cite{lin2024preflmr} spans both textual and multimodal resources, while WebQA~\cite{chang2022webqa} requires retrieval over heterogeneous sources of images or text, often with multi-hop queries.
\Cref{tab:benchmark_comparison}, compares {\datasetname} with existing MKB-VQA benchmarks.
Prior benchmarks~\cite{shahMYP19, 10.1145/3477495.3531753, chen2023can, mensink2023encyclopedic} pair each question with an image of the target document's \emph{main entity}, introducing the visual shortcut.
As shown in \cref{tab:benchmark_comparison} {\datasetname} is the only benchmark whose query images depict \emph{related} entities, requiring retrievers to jointly reason over visual and textual signals.

\noindent \textbf{Multimodal Retrievers} 
Foundational multimodal retrievers such as CLIP~\cite{radford2021learning}, Image-to-Word~\cite{Mori1999ImagetowordTB}, and BiT~\cite{kolesnikov2020bigtransferbitgeneral} align visual and textual representations in a shared embedding space for cross-modal retrieval. To handle queries combining text and images, this paradigm has been extended by ReVIVE~\cite{lin2022revive}, FLMR~\cite{lin2023finegrainedlateinteractionmultimodalretrieval}, and MuKA~\cite{deng-etal-2025-muka}, among other joint-modality architectures~\cite{Gao_2022_CVPR, long2025retrieval}. Universal frameworks—UniIR~\cite{wei2024uniir}, CAFE~\cite{yu2025cafeunifyingrepresentationgeneration}, MMEmbed~\cite{lin2025mmembeduniversalmultimodalretrieval}, GENIUS~\cite{kim2025genius}, LAMRA~\cite{liu2025lamra}, and PreFLMR~\cite{lin2024preflmr}—further support arbitrary query and document modalities. Despite these advances, existing retrievers encode each document with a single image, limiting retrieval on {\datasetname}, where queries involve diverse related entities. {\modelname} addresses this via multi-image encoding, yielding robust representations.

\added{\noindent \textbf{Multimodal Knowledge Graph RAG} \ 
Another line of work structures the knowledge base itself as a multimodal knowledge graph. RAG-Anything~\cite{guo2025raganything} builds cross-modal graphs over documents with hybrid graph and vector retrieval, and mKG-RAG~\cite{yuan2025mkgrag} constructs modality-aligned entity graphs for knowledge-intensive VQA with a dual-stage document-then-graph retrieval. These are full RAG pipelines that retrieve graph nodes and subgraphs for a generator, whereas {\modelname} is a document retriever that keeps the KB unchanged and only enriches each document's representation with images of its related entities, without typed relations or graph reasoning. We compare against both under a shared protocol in \cref{app:kgrag}.}

\section{Dataset Construction}
\label{sec:pipeline}
We introduce {\datasetname}, a benchmark addressing the visual shortcut bias in existing MKB-VQA benchmarks. Unlike prior benchmarkss, queries in {\datasetname} cannot be solved through visual shortcuts, providing a balanced assessment when combined with prior benchmarks. For scalability, we develop an automated LLM-driven generation pipeline (\Cref{fig:pipeline}), yielding 120K training set and a 2K human-curated test set.

\subsection{MKB-VQA Sample Generation Pipeline}
While LLMs exhibit strong reasoning capabilities, their instruction-following degrades over long contexts such as Wikipedia~\cite{liu2023lostmiddlelanguagemodels}. We overcome this with a graph-based approach that ensures controllable generation, essential for maintaining the intended benchmark structure. Specifically, our pipeline comprises three stages: (a) converting the document into a knowledge graph, (b) selecting target subgraphs, and (c) generating multimodal QA from selected subgraphs.

\noindent{\textbf{Knowledge Graph Construction}} \ \
Wikipedia documents center on a primary subject, which we call the main entity. They typically contain an image of this entity, but using it as the query image introduces a visual shortcut. To circumvent this, we construct a one-hop graph centered on the main entity to identify related entities within the document, then generate queries paired with images of these related entities instead—keeping the benchmark shortcut-free.
To extract these entities, we prompt an LLM to identify named entities that co-occur with the main entity and bear a semantic relationship to it. Importantly, we do not fully convert the document into a structured entity–relation graph, as this often discards contextual information in the natural language—causal descriptions, attributes, temporal dependencies, and explanatory qualifiers. Instead, each edge is annotated with a natural language sentence describing the entity's relationship to the main entity. This representation enables precise control over target-entity selection for QA generation while preserving the original descriptive detail. \Cref{fig:pipeline}a illustrates this process: for the Wikipedia page on `Pennine Cycleway', the LLM identifies related entities such as `Scammonden Reservoir' and `Cumbria county' as neighboring nodes in the one-hop graph.

\noindent{\textbf{Target Subgraph Extraction}} \ \ 
MKB-VQA samples involve two related concepts: one depicted by the query image and one serving as the answer. 
Grounding QA in the Wikipedia KB, we instantiate these concepts as two entities per document: the main entity and a randomly selected related entity.
Since using the main entity as the query image would reintroduce the visual shortcut, we assign it as the answer entity and the related entity as the query entity.
Each pair, together with the tagged sentence describing their relation, forms the basis of an MKB-VQA sample.
However, questions formulated directly from this pair can be ambiguous, with multiple entities validly satisfying the question.
For example, in \Cref{fig:pipeline}b, the relation `The Pennine Cycleway crosses a dam at Scammonden Reservoir' with a query image of the reservoir may yield `Which cycling path traverses the dam of this reservoir?'---which admits multiple correct answers.
Since fully removing such ambiguity is difficult in an automated pipeline, we reduce it by introducing a qualifying entity: an additional randomly selected related entity that provides contextual cues to narrow down the answer.
The final target subgraph---query, answer, and qualifying entities---is passed to the LLM to generate the QA.

\noindent{\textbf{LLM-based VQA Sample Generation}} \ \ 
Given a target subgraph, we generate a QA sample comprising a query image, query text, and answer text.
We first retrieve an image of the query entity from M2KR~\cite{lin2024preflmr} as the query image.
We then prompt the LLM to formulate a question answered by the answer entity, conditioned on the two relation descriptions in the subgraph.
To ensure the query image is indispensable, the LLM is instructed to refer to the query entity not by name but with a demonstrative phrase (\eg, ``this organization'' or ``this monument'').
Finally, the question is paraphrased by the LLM, and we discard questions whose source document is retrieved by BM25~\cite{robertson1995okapi} in the top-5 results, reducing lexical overlap.
Full prompts are in the Appendix; \Cref{fig:pipeline}c illustrates the process with an example QA pair.

\subsection{{\datasetname} Bench Construction}
Following this procedure, we construct {\datasetname} using Qwen-2.5-32B-Instruct~\cite{bai2023qwen}, comprising a 120K training set and a 2K test set.
For the test set, we additionally apply a strict human selection stage that retains queries with an unambiguous answer that require both modalities to solve---a stricter bar than prior MKB-VQA benchmarks apply, ensuring automatic evaluation is reliable (\eg, a question such as ``What is the name of the plant that can be found in this state as well as in Florida?'' admits multiple valid answers and is excluded for evaluation; see \cref{app:curation}).
\added{We distinguish \emph{factual validity} from \emph{evaluation strictness}: a human audit finds 98.4\% of generated samples factually valid, but only 29.16\% also satisfy answer uniqueness and multimodal necessity. Rejected samples are mostly correct questions with multiple valid answers (75.35\%) or ones solvable without the image (21.27\%); they remain valid retrieval supervision and are kept for training (\cref{app:curation,app:train_quality}).
Accordingly, the shortcut-free property holds for both splits by construction, since every query image depicts a related rather than the main entity, while the test set is additionally verified by humans to require both modalities.}
We divide the test set into a seen subset, whose GT documents overlap with the training data, and an unseen subset without such overlap.
Finally, to guarantee a single GT document per query, we remove related entities whose documents reciprocally reference the main entity, preventing cases where both documents would reveal the answer.

\begin{figure*}[t]
    \centering  
    \includegraphics[width=1.0\linewidth]{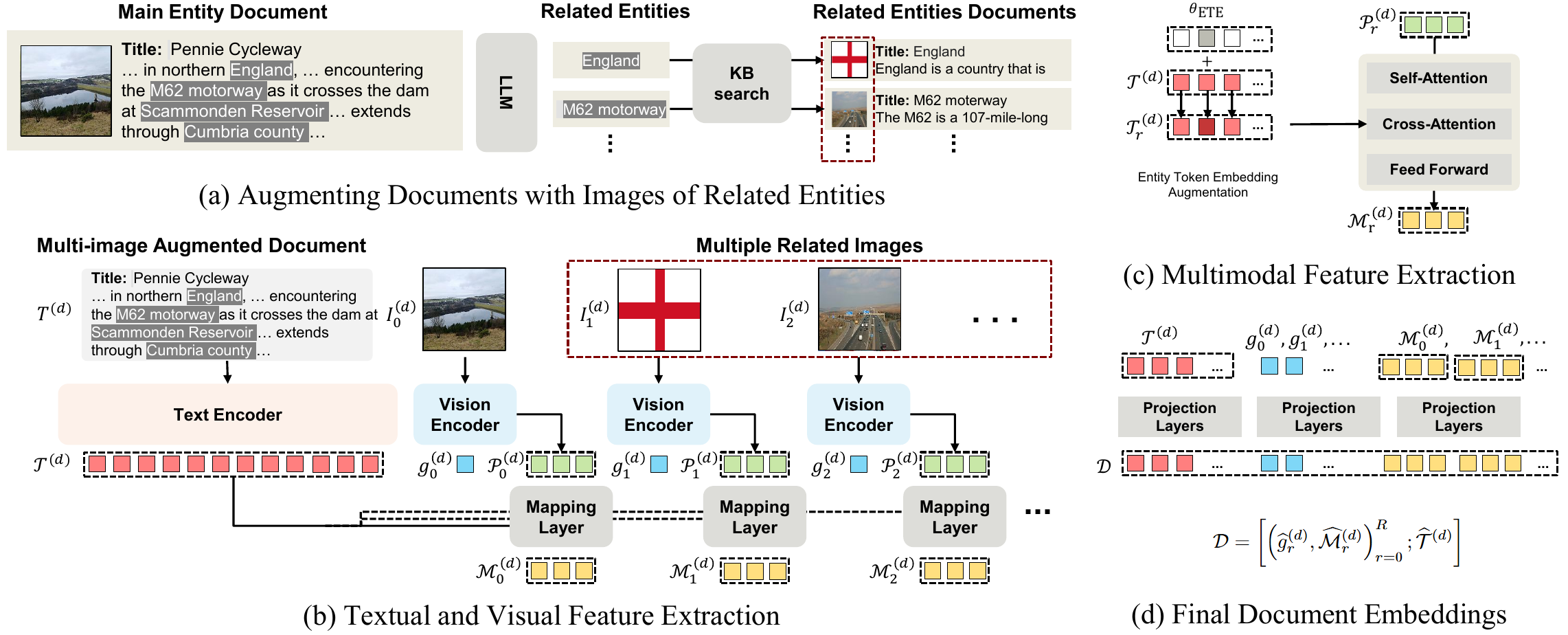}
    \caption{
    \textbf{Overview of {\modelname} Document Encoder Architecture.}
     (a) given a document, related named entities are identified with an LLM~\cite{bai2023qwen}, and corresponding images are collected from the KB; (b) textual, global image, and patch-level features are extracted, with patch features attending to textual features through cross-attention to yield multimodal features; and (c) entity token embeddings are incorporated into the textual features prior to cross-attention for richer contextualization; (d) the final document embedding jointly integrates textual, global, and multimodal features projected into the same embedding space.
    }
    \label{fig:main}
\end{figure*}

\section{Method}
\label{sec:method}
Existing multimodal retrievers~\cite{deng-etal-2025-muka,caffagni2025recurrence,wei2024uniir} typically encode each document with a single image depicting its main entity.
While effective on existing MKB-VQA benchmarks, they falter on {\datasetname}: since document embeddings carry only the main entity's image, retrieval often fails when the query image depicts a non-main entity of the target document.
To address this, we introduce {\modelname}, a multimodal retriever whose core mechanism enriches document representations with images of multiple related entities, plus two lightweight, near-parameter-free refinements detailed below.
{\modelname} builds upon MuKA~\cite{deng-etal-2025-muka}, a single-image document retriever, extending its representation capacity.
Below, we review MuKA's scoring mechanism and query encoder (\cref{sec:late_interaction,sec:query_emb}), then detail our extended document embedding module (\cref{sec:doc_emb}).

\subsection{Late Interaction Scoring}
\label{sec:late_interaction}
Following the ColBERT family of multimodal retrievers~\cite{deng-etal-2025-muka, lin2024preflmr}, we represent the query and document as sets of token vectors, $\mathcal{Q} = \{ q_1, \dots, q_m \}$ and $\mathcal{D} = \{ d_1, \dots, d_n \}$.
Each query token contributes its maximum similarity over all document tokens, and these contributions are summed to yield the relevance score:
\begin{equation}
s(\mathcal{Q}, \mathcal{D}) = \sum_{i=1}^{m} \max_{j \in [1, n]} \text{sim}(q_i, d_j),
\end{equation}
where $\text{sim}(\cdot,\cdot)$ denotes cosine similarity.
Despite the token-wise computation, efficient indexing~\cite{santhanam2022plaidefficientenginelate} keeps retrieval practical.
This token-level formulation lets visual and textual query tokens match independently—visual tokens against similar document images, textual tokens against relevant text.

\subsection{Query Encoder}
\label{sec:query_emb}
The query encoder comprises image and text encoders and a cross-attention transformer block, each paired with dedicated projection layers.
Given a query $(I^{(q)}, T^{(q)})$, the goal is to produce the query feature set $\mathcal{Q}$ for scoring.
A ColBERTv2 text encoder~\cite{santhanam2022colbertv2effectiveefficientretrieval} yields $N_t$ token-level text embeddings $\mathcal{T}^{(q)} = \{t^{(q)}_{1}, \dots, t^{(q)}_{N_t}\}$, while a CLIP image encoder~\cite{radford2021learning} yields a global image feature $g^{(q)}$ and, from its penultimate layer, $N_p$ patch-level features $\mathcal{P}^{(q)} = \{p^{(q)}_{1}, \dots, p^{(q)}_{N_p}\}$.
The patch features are contextualized with the text via cross-attention~\cite{vaswani2023attentionneed}, using $\mathcal{P}^{(q)}$ as queries and $\mathcal{T}^{(q)}$ as keys and values, producing $N_p$ multimodal features $\mathcal{M}^{(q)} = \{m^{(q)}_{1}, \dots, m^{(q)}_{N_p}\}$.
Finally, $g^{(q)}$, $\mathcal{T}^{(q)}$, and $\mathcal{M}^{(q)}$ are projected into $\mathbb{R}^{D}$ by their dedicated projection layers (denoted $\widehat{\cdot}$), and concatenated to form the query feature set $\mathcal{Q} = [\widehat{g}^{(q)}; \widehat{\mathcal{T}}^{(q)}; \widehat{\mathcal{M}}^{(q)}]$.

\subsection{Document Encoder}
\label{sec:doc_emb}
\Cref{fig:main} overviews our document encoder.
Each document consists of a text and a main-entity image, following~\citet{deng-etal-2025-muka}.
We (a) augment it with images of related entities collected from the KB, (b) encode the text and all images into textual and visual features, (c) fuse each image with its relevant text via a multimodal mapping layer, where an Entity Token Embedding (ETE) marks the tokens of the entity depicted in each image, and (d) concatenate all features into the document representation $\mathcal{D}$.
We describe each step below.

\noindent{\textbf{(a) Augmenting Documents with Related Entity Images}} \ \
Given the textual content $T^{(d)}$, we identify all named entities related to the main entity with Qwen2.5-32B~\cite{bai2023qwen}, query the KB for each entity's document, and use its primary image as the entity's visual representation.
The augmented document is $(I^{(d)}_0, I^{(d)}_1, \dots, I^{(d)}_R, T^{(d)})$, where $I^{(d)}_0$ depicts the main entity and $I^{(d)}_r$, $r \in \{1,\dots,R\}$, depict related entities.
Note that we use an LLM only because the Wikipedia passages in M2KR~\cite{lin2024preflmr} are distributed without hyperlink label; entities are subsequently filtered to those with a corresponding KB document, so the resulting set approximates the Wikipedia's hyperlink structure. 
In deployment, hyperlinks can be used directly without an LLM.
\added{The source of the links (LLM extraction or hyperlinks) is interchangeable; what is essential is that each added image depicts an entity that the document actually relates to. Images of unrelated entities or additional views of the main entity, even at the same index size, do not yield the gain (\cref{sec:capacity}).}
\\
\noindent{\textbf{(b) Textual and Visual Feature Extraction}} \ \
Using the same ColBERTv2 and CLIP encoders as the query encoder, we extract token-level textual features $\mathcal{T}^{(d)} = \{t^{(d)}_{1}, \dots, t^{(d)}_{N_t}\}$ from $T^{(d)}$, and a global feature $g^{(d)}_r$ and patch-level features $\mathcal{P}^{(d)}_r = \{p^{(d)}_{r,1}, \dots, p^{(d)}_{r,N_p}\}$ from each image $I^{(d)}_r$, $r \in \{0,\dots,R\}$.
\begin{table*}[t]
\centering
\caption{
\textbf{Evaluating Robustness Against Visual Shortcuts}. Retrieval and QA performance are reported. All models are jointly trained on EVQA and {\datasetname}. The GT Document row represents an upper bound using gold knowledge. 
{\modelname} consistently outperforms prior methods in both settings with and without visual shortcuts, showing the benefit of leveraging multi-image context over single-image or text-only approaches.
}
\scalebox{0.75}{
\begin{tabular}{llcccccccc}
\toprule
& & \multicolumn{2}{c}{\textbf{InfoSeek}} & \multicolumn{2}{c}{\textbf{EVQA}} & \multicolumn{4}{c}{\textbf{{\datasetname}}} \\
\cmidrule[.01pt](lr){3-4} \cmidrule[.01pt](lr){5-6} \cmidrule[.01pt](lr){7-10}
 & & & & & & \multicolumn{2}{c}{Seen} & \multicolumn{2}{c}{Unseen} \\
\cmidrule[.01pt](lr){7-8} \cmidrule[.01pt](lr){9-10}
\textbf{Method} & \textbf{Document Modality} & Recall@5 & Acc & Recall@5 & BEM & Recall@5 & BEM & Recall@5 & BEM \\
\specialrule{1pt}{2pt}{2pt}
PreFLMR~\cite{lin2024preflmr} & text & 40.7 & 33.5 & 63.9 & 57.0 & \textcolor{black}{23.6} & 19.9 & 11.1 & 13.6 \\
UniIR~\cite{wei2024uniir} & text + single-image & 34.5 & 26.4 & 22.0 & 25.7 & \textcolor{black}{13.8} & 14.6 & 4.7 & 11.2 \\
ReT~\cite{caffagni2025recurrence} & text + single-image & 50.1 & 40.4 & 46.3 & 43.7 & \textcolor{black}{15.1} & 15.2 & 6.8 & 11.4 \\
MuKA~\cite{deng-etal-2025-muka} & text + single-image & 51.7 & 41.0 & 64.9 & 60.7 & \textcolor{black}{20.8} & 18.7 & 8.6 & 12.2 \\
\textbf{{\modelname} (Ours)} & \textbf{text + multi-image} & \textbf{53.3} & \textbf{43.2} & \textbf{65.8} & \textbf{62.2} & \textbf{43.2} & \textbf{32.7} & \textbf{36.9} & \textbf{31.6} \\
\cmidrule(lr){1-10}
\textcolor{gray}{GT Document} & \textcolor{gray}{--} & \textcolor{gray}{--} & \textcolor{gray}{76.5} & \textcolor{gray}{--} & \textcolor{gray}{88.2} & \textcolor{gray}{--} & \textcolor{gray}{68.5} & \textcolor{gray}{--} & \textcolor{gray}{70.2} \\
\bottomrule
\end{tabular}
}
\label{tab:main}
\end{table*}
\\
\noindent{\textbf{(c) Entity-Aware Multimodal Feature Extraction}} \ \
To capture information available through joint modeling of both modalities, each image's patches are fused with the text by a multimodal mapping layer---a cross-attention transformer block.
Since each image corresponds to a known named entity in $T^{(d)}$, we expose this association to the fusion rather than leaving alignment entirely to implicit attention: an Entity Token Embedding $\theta_{\text{ETE}}$, a learnable vector, is added to the entity's tokens, forming an image-specific textual set $\mathcal{T}^{(d)}_r$ via $t^{(d)}_{s} \leftarrow t^{(d)}_s + \theta_{\text{ETE}}$ for all $s \in \mathcal{S}_r$, where $\mathcal{S}_r$ denotes the token indices of the entity of $I^{(d)}_r$ (\textcolor{gray}{gray} tokens in \cref{fig:main}c).
The mapping layer then attends from $\mathcal{P}^{(d)}_r$ (query) to $\mathcal{T}^{(d)}_r$ (key and value), producing multimodal features $\mathcal{M}^{(d)}_r = \{m^{(d)}_{r,1}, \dots, m^{(d)}_{r,N_p}\}$ that are explicitly grounded in each image's entity.
\\
\noindent{\textbf{(d) Final Document embeddings}} \ \
All features are projected into $\mathbb{R}^D$ and concatenated:
\begin{equation}
    \mathcal{D} = \left[ \left( \widehat{g}^{(d)}_r, \widehat{\mathcal{M}}^{(d)}_r \right)_{r=0}^R; \widehat{\mathcal{T}}^{(d)} \right].
\end{equation}
Under late interaction (\cref{sec:late_interaction}), a query image depicting any related entity can now match its corresponding $\widehat{g}^{(d)}_r$ or $\widehat{\mathcal{M}}^{(d)}_r$, enabling retrieval without the visual shortcut.
\\
\noindent{\textbf{Relation to MuKA}} \ \
{\modelname} extends MuKA, which encodes only $(I^{(d)}_0, T^{(d)})$ and Concatenates patches with the full, unmarked text.
Our extension adds no learnable parameters beyond $\theta_{\text{ETE}}$, isolating the gains to the enriched representation. \added{We verify in \cref{sec:capacity} that these gains are not explained by the larger document representation alone.}

\section{Experiments}
\label{sec:experiments}
\subsection{Experimental Setup}
\label{sec:experimental_setup}
\noindent\textbf{Datasets} \ \ 
We evaluate on {\datasetname} and two existing benchmarks, EVQA~\cite{mensink2023encyclopedic} and InfoSeek~\cite{chen2023can}.
Since {\datasetname} excludes visual shortcuts while EVQA and InfoSeek contain them, their combination enables comprehensive assessment of realistic MKB-VQA scenarios.
{\datasetname} contains 120K training samples and a 2K human-curated test set; EVQA provides 167K training and 3.7K test samples; InfoSeek consists of 676K training samples---of which we sample 100K following~\citet{lin2024preflmr}---and 4.7K test samples.
All three knowledge bases consist of textual Wikipedia documents (\cref{app:kb}), augmented with a main-entity image following~\citet{deng-etal-2025-muka} and further enriched with related-entity images as described in \cref{sec:doc_emb}.

\noindent\textbf{Metrics} \ \
We evaluate both retrieval and answer generation.
For retrieval, we report Recall@$K$, the proportion of queries whose annotated document appears in the top-$K$ results.
For generation, we use BERT Matching (BEM)~\cite{bulian-etal-2022-tomayto} on {\datasetname} and EVQA---a soft semantic measure beyond strict lexical matching---and InfoSeek's official Accuracy metric with answer-type-specific matching functions~\cite{chen2023can}.
Unless otherwise specified, all retrievers are evaluated by feeding their top-5 results to the same VILA-13B~\cite{liu2023visualinstructiontuning}, following~\citet{deng-etal-2025-muka}.

\noindent\textbf{Models} \ \
We compare {\modelname} against four multimodal baselines: UniIR~\cite{wei2024uniir}, ReT~\cite{caffagni2025recurrence}, PreFLMR~\cite{lin2024preflmr}, and MuKA~\cite{deng-etal-2025-muka}.
{\modelname}, PreFLMR, and MuKA share the same core encoders (ColBERT-base-v2 for text, CLIP ViT-G for vision) and differ mainly in the document encoder: PreFLMR uses text-only documents, MuKA attaches a single main-entity image, and {\modelname} incorporates multiple related-entity images (\cref{sec:doc_emb}).
Implementation details are in \cref{app:details}.

\subsection{Results}
\noindent\textbf{Evaluating Robustness Against Visual Shortcuts} \ \
We evaluate retrievers on {\datasetname}, InfoSeek, and EVQA, finetuning all models on the combined EVQA and {\datasetname} training sets to cover settings both with and without visual shortcuts.
As shown in \cref{tab:main}, single-image multimodal retrievers struggle on RETINA—even falling behind text-only baselines. 
This reversal highlights the effect of visual shortcuts: single-image models overfit to shortcut cues. 
Yet text-only recall itself remains low (23.6/11.1), confirming neither modality alone suffices on RETINA.
By augmenting documents with multiple related-entity images, {\modelname} achieves nearly twice the recall of the best baseline on {\datasetname} while remaining robust on existing benchmarks.
We additionally report performance with GT documents as an upper bound.

\begin{table}[t]
    \centering
    \setlength{\tabcolsep}{4pt}
    \scalebox{0.8}{
        \begin{tabular}{llcccc}
        \toprule
        & & \textbf{InfoSeek} & \textbf{EVQA} & \multicolumn{2}{c}{\textbf{{\datasetname}}} \\
        \cmidrule(lr){3-3} \cmidrule(lr){4-4} \cmidrule(lr){5-6}
        \textbf{Train} & \textbf{Method} & & & Seen & Unseen \\
        & & R@5 & R@5 & R@5 & R@5 \\
        \midrule
        \multirow{3}{*}{\textbf{InfoSeek}}
        & PreFLMR & 43.3 & 62.4 & 18.5 & 7.9 \\
        & MuKA    & \textbf{54.0} & 63.8 & 17.4 & 7.7 \\
        & \textbf{{\modelname}} & 53.9 & \textbf{64.3} & \textbf{35.2} & \textbf{32.7} \\
        \midrule
        \multirow{3}{*}{\textbf{EVQA}}
        & PreFLMR & 41.2 & 64.4 & 19.9 & 9.0 \\
        & MuKA    & 51.9 & 65.8 & 18.3 & 8.3 \\
        & \textbf{{\modelname}} & \textbf{53.5} & \textbf{66.0} & \textbf{39.9} & \textbf{34.0} \\
        \midrule
        \multirow{3}{*}{\textbf{{\datasetname}}}
        & PreFLMR & 31.8 & 53.6 & 24.6 & 13.0 \\
        & MuKA    & 42.1 & 56.5 & 22.2 & 10.1 \\
        & \textbf{{\modelname}} & \textbf{46.3} & \textbf{58.7} & \textbf{44.0} & \textbf{37.6} \\
        \bottomrule
        \end{tabular}
    }
    \caption{\textbf{Cross-dataset Transfer.} Rows indicate the training source, columns the evaluation target.}
    \label{tab:transfer}
\end{table}

\noindent\textbf{Cross-Dataset Transfer} \ \
\Cref{tab:transfer} reports cross-dataset transfer evaluations across InfoSeek, EVQA, and {\datasetname}. 
Baselines trained on InfoSeek or EVQA fail to generalize when evaluated on the shortcut-free setting of {\datasetname}, dropping below 10\% R@5 on unseen splits. 
In contrast, {\modelname} demonstrates strong cross-dataset generalization; even when trained exclusively on InfoSeek or EVQA, it achieves 32.7\% and 34.0\% R@5 on {\datasetname} Unseen, respectively. 
Furthermore, when trained on {\datasetname}, {\modelname} retains competitive performance when transferred back to InfoSeek (46.3\% R@5) and EVQA (58.7\% R@5). 
While a moderate performance drop is observed compared to in-domain training, {\modelname} avoids catastrophic performance degradation, showing that {\datasetname} provides generalizable supervision without overfitting to shortcut cues.

\begin{figure}[t]
    \centering
    \includegraphics[width=\linewidth]{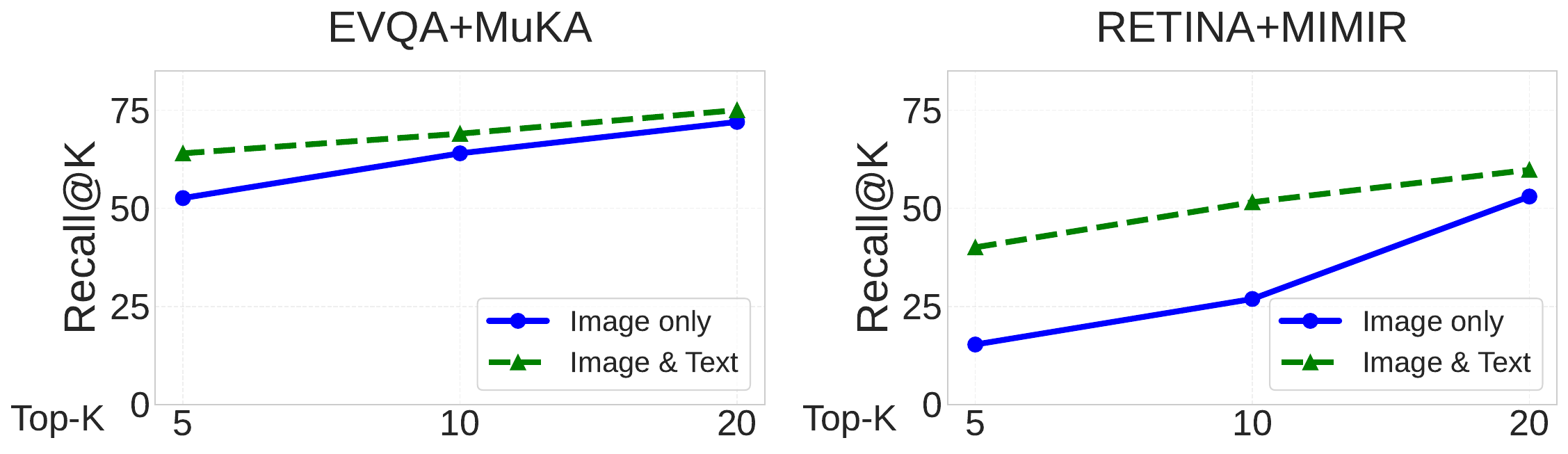}
    \caption{\textbf{Image-only vs. Multimodal retrieval.} On EVQA (left), image-only MuKA (\textcolor{blue}{blue}) nearly matches the multimodal model (\textcolor{green!60!black}{green}), revealing the visual shortcut. On {\datasetname} (right), {\modelname} shows a substantial gap, indicating that multimodal processing is required.}
    \label{fig:shortcut_clarification}
\end{figure}

\noindent\textbf{The Visual Shortcut} \ \
Existing benchmarks pair the query image with the GT document's main entity, so a dense visual signal can dominate the sparse textual match (\cref{sec:late_interaction}) and recover the target on its own (\cref{sec:intro}). {\datasetname} breaks this coupling by letting the query depict any related entity, and \cref{fig:shortcut_clarification} shows the consequence: on EVQA (left), image-only MuKA (\textcolor{blue}{blue}) retains nearly 90\% (\textcolor{green!60!black}{green}) at top-k cutoffs (e.g., Recall@5 and @10). On {\datasetname} (right), however, the image-only performance drops by nearly half. 

\begin{figure*}[t]
  \centering
    \includegraphics[width=\linewidth]{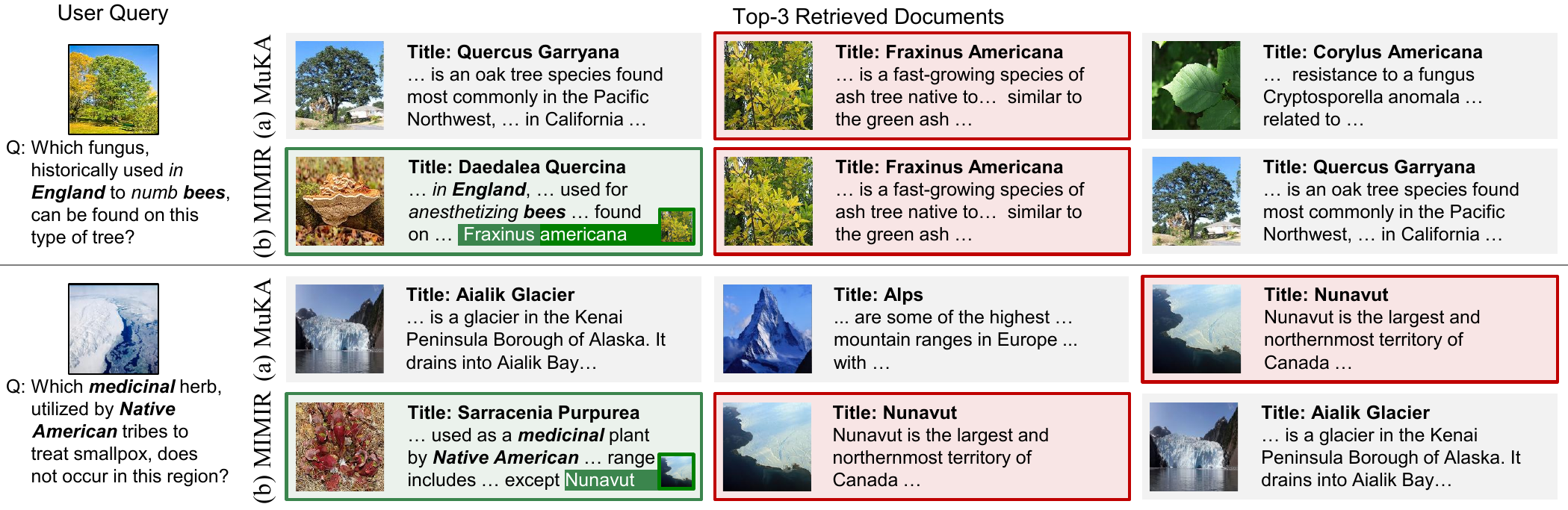}
  \caption{
  \textbf{Qualitative Comparison of Retrieval on {\datasetname} Benchmark.} Results for (a) MuKA~\cite{deng-etal-2025-muka} and (b) {\modelname}. GT documents are in \textbf{\textcolor{figgreen}{green}}; hard distractors containing the query image entity are in \textbf{\textcolor{figbrightred}{red}}; query entity in GT documents are highlighted in \textbf{\textcolor{figgreen}{green}}. 
  While MuKA is biased toward visual shortcuts and often retrieves documents containing images similar to the query, {\modelname} leverages related images 
  \protect\raisebox{-0.4ex}{\includegraphics[height=2ex]{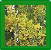}} (Fraxinus americana) and \protect\raisebox{-0.4ex}{\includegraphics[height=2ex]{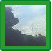}} (Nunavut), which are augmented to the document embeddings.%
  }
  \label{fig:examples}
\end{figure*}

\added{\noindent\textbf{Robustness to Visual-similar Distractors}} \ \ \added{Against hard distractors---non-GT documents whose main entity matches the query image---MuKA retrieves distractors more often than GT documents at Recall@1, whereas {\modelname} consistently ranks GT documents higher, showing that it relies on text as well as visual similarity (\cref{app:distractor}).}

\added{\noindent\textbf{Disentangling Capacity from Entity Association}}\label{sec:capacity} \ \ \added{Since {\modelname} indexes more vectors per document than single-image baselines, we test whether its gain stems from added capacity or from \emph{which} evidence is indexed (\cref{tab:capacity}).
Each control enlarges the document representation to {\modelname}'s index size: (i) +repr.\ adds indexed vectors; (ii) +augmented images attaches as many crops/flips of the main-entity image as {\modelname} adds images; (iii) +related passages appends to PreFLMR the passages of the \emph{same} related entities, in text instead of images.
(i) and (ii) do not help on {\datasetname}, so capacity alone does not explain the gain.
(iii) gains little on {\datasetname} while sharply degrading InfoSeek and EVQA: {\datasetname} queries refer to the entity only through the image, and the appended text turns non-GT documents into distractors.
Conversely, with the number of images fixed, replacing related-entity images with unrelated ones monotonically lowers R@5 (43.2 $\rightarrow$ 36.8 at 0$\rightarrow$30\%; \cref{tab:selection_noise}b).
The gain thus comes from indexing the appearance of the document's related entities, not from a larger representation.}

\begin{table}[t]
    \centering
    \revcolor
    \setlength{\tabcolsep}{3.5pt}
    \resizebox{\linewidth}{!}{
        \begin{tabular}{llcccc}
        \toprule
        & & \textbf{InfoSeek} & \textbf{EVQA} & \multicolumn{2}{c}{\textbf{{\datasetname}}} \\
        \cmidrule(lr){5-6}
        \textbf{Method} & \textbf{Size-matched aug.} & & & Seen & Unseen \\
        \midrule
        PreFLMR & -- & 40.7 & 63.9 & 23.6 & 11.1 \\
        PreFLMR & + repr. & 41.0 & 64.2 & 24.2 & 12.4 \\
        PreFLMR & + related passages & 27.3 & 43.1 & 27.2 & 14.7 \\
        \midrule
        MuKA & -- & 51.7 & 64.9 & 20.8 & 8.6 \\
        MuKA & + repr. & 52.0 & 64.8 & 21.2 & 8.9 \\
        MuKA & + augmented images & 52.9 & 65.3 & 17.2 & 6.3 \\
        \midrule
        \textbf{{\modelname}} & -- & \textbf{53.3} & \textbf{65.8} & \textbf{43.2} & \textbf{36.9} \\
        \bottomrule
        \end{tabular}
    }
    \caption{\added{\textbf{Capacity-Controlled Comparison} (R@5). Each control enlarges the per-document representation to {\modelname}'s index size without adding related-entity visual content.}}
    \label{tab:capacity}
\end{table}

\added{\noindent\textbf{Efficiency and Index Size}} \ \ \added{Related-entity images increase indexed tokens by 2.3$\times$, a one-time offline cost, but search cost does not grow proportionally: PLAID~\cite{santhanam2022plaidefficientenginelate} prunes candidates via centroid interaction and fully scores only a small candidate set. Latency is dominated by query encoding ($\sim$2\,s) rather than retrieval ($\sim$0.002\,s), and the query encoder is identical to that of the single-image baseline (\cref{app:efficiency}). $K$, the number of related-entity images per document, controls index growth: even $K{=}1$ surpasses MuKA (24.7 vs.\ 20.8 R@5 on Seen; \cref{tab:selection_noise}a), so $K$ can be capped to trade index size for recall.}

\noindent\textbf{Qualitative Analysis} \ \
\Cref{fig:examples} compares MuKA and {\modelname} on {\datasetname}.
While MuKA also account for text, they tend to be heavily dominated by visual similarity, leading to exploit shortcut biases (\eg, trees or snowy landscapes in \cref{fig:examples}a) that push the GT document down despite its text relevance (\eg, \textit{medicinal} in \cref{fig:examples}b).
In contrast, {\modelname} identifies related entities during document encoding---\eg, Fraxinus americana' and Nunavut' in \cref{fig:examples}b---and integrates their visual features into the document representation.
This integration explicitly boosts the GT document's visual similarity score, which would otherwise remain low, overshadowing the textual similarity.
Consequently, {\modelname} achieves a higher rank for the GT document through balanced visual and textual alignment.

\begin{table}[t]
\centering
\caption{
\textbf{{\modelname} Component Ablation.} 
Cumulative impact of components on retrieval performance: baseline, +MI (Multiple Images), +MMF (Multimodal Features), and +ETE (Entity Token Embedding).
}
\label{tab:03_ablation_tab}
\scalebox{0.65}{%
\begin{tabular}{l c c c c}
\toprule
 & \textbf{InfoSeek} & \textbf{EVQA} & \multicolumn{2}{c}{\textbf{\datasetname}} \\
\cmidrule(lr){4-5}
\textbf{Method} &  &  & Seen & Unseen \\
\specialrule{1pt}{2pt}{2pt}
Baseline (MuKA \cite{deng-etal-2025-muka}) & 51.7 & 64.9 & 20.8 & 8.6 \\ 
\quad + MI & 51.5 & 65.1 & 38.1 & 33.3 \\
\quad + MMF & 51.6 & 65.3 & 41.1 & 34.5 \\
\quad + ETE (Full {\modelname}) & 53.3 & 65.8 & 43.2 & 36.9 \\ 
\bottomrule
\end{tabular}
}
\end{table}

\noindent{\textbf{{\modelname} Component Ablation}} \ \
\cref{tab:03_ablation_tab} ablates multiple images (MI), multimodal features (MMF), and entity token embeddings (ETE) from \cref{sec:doc_emb}.
The baseline corresponds to the original MuKA; thus +MI directly measures multi-image augmentation on the strongest single-image baseline.
% [Original] Adding multiple images (+MI)—here simply concatenating the projected global features of the additional images, as MuKA lacks multimodal document features—accounts for most of the RETINA gain (seen: 20.8 → 38.1), the essential ingredient for shortcut-free retrieval.
\added{Adding the images of the related entities identified in \cref{sec:doc_emb}(a) (+MI)---whose projected global features are concatenated to the document representation, as MuKA lacks multimodal document features---accounts for most of the {\datasetname} gain (seen: 20.8 $\rightarrow$ 38.1). As shown in \cref{sec:capacity}, this gain comes from the entity association of the added images, not from their number.}
Furthermore, +MMF and +ETE add ~2 points at nearly zero parameter cost, the former by enriching the document representation and the latter by aligning cross-attention with image-relevant text tokens.

\section{Conclusion}
\label{sec:conclusion}
We propose {\datasetname}, a novel MKB-VQA benchmark that eliminates the visual shortcut present in prior datasets. We show that existing models exploit this shortcut, limiting their practical utility. 
{\datasetname} features a human-curated test set and large-scale training data generated via automatic graph-based pipeline. 
We further introduce {\modelname}, a framework designed for retrieval without over-relying on visual shortcuts. \added{Capacity-controlled analyses show that its gains come from indexing images of each document's related entities rather than from a larger representation.} Extending to other modalities remains a promising direction.

\section*{Limitations}
First, {\modelname} requires a higher indexing cost than single-image retrievers, as encoding the images of related entities increases offline processing time and index size. While per-query latency is largely unaffected, this storage overhead may become a burden for very large knowledge bases.
Second, {\datasetname} relies on an LLM for entity extraction and question generation, which may introduce model-specific biases. Only the test set is human-verified, so the training set may still contain ambiguous questions.
Finally, both {\datasetname} and {\modelname} are built on English Wikipedia, and their generalization to other domains, languages, and knowledge bases remains untested.
\section*{Acknowledgements}
\added{This research was supported by MSIT (IITP: IITP-2026-RS-2024-00436857, RS-2024-00398115, IITP-2026-RS-2020-II201819, IITP-2026-RS-2025-02304828, RS-2026-25507282, RS-2026-25585074; NRF: RS-2025-23523979), MCST (KOCCA: RS-2026-25506607, RS-2024-00345025), MND (RS-2026-25551943) and MSS (RS-2026-25549378). This research was also supported by the ``Advanced GPU Utilization Support Program'' funded by the Government of the Republic of Korea (MSIT), and by a research grant from Google.}

\bibliography{main}

\appendix
\label{sec:appendix}
\clearpage
\setcounter{page}{1}
\title{Breaking the Visual Shortcuts in Multimodal Knowledge-Based Visual Question Answering} 
\maketitle
\appendix

\section{{\datasetname} Bench Construction}
This section describes the process of constructing {\datasetname} bench, including how textual queries and images were collected, processed, and curated to ensure quality. Throughout all stages of this pipeline, Qwen2.5-32B~\cite{bai2023qwen} was used.

\subsection{Graph-Guided Generation under Compositional Constraints}
\label{app:graph_guided}
Generating a shortcut-free MKB-VQA sample requires satisfying several constraints at once: the question must reference the query entity indirectly, resolve to a unique answer entity, and remain grounded in the document's factual relations.
Prompting an LLM to produce such a sample directly from a full Wikipedia document is unreliable, for two well-documented reasons: instruction-following degrades as constraints compose~\cite{wen2024benchmarking}, and reasoning degrades over long input contexts~\cite{liu2023lostmiddlelanguagemodels}.
Rather than ask the model to juggle these demands in a single pass, we decompose the task with an intermediate graph representation~\cite{ainslie2023gqa, tran2025reasonvqa, trivedi2022musiquemultihopquestionssinglehop, reddy-etal-2017-generating}.
By first extracting a one-hop subgraph and then conditioning generation on a small, explicit skeleton---an answer entity, a query entity, a qualifying entity, and their relation sentences---we reduce each generation step to a bounded, controllable operation, which is what keeps the intended benchmark structure intact at scale.
Because every entity in the skeleton is required to have a corresponding document in our knowledge base, the LLM operates as a linguistic synthesizer over pre-verified entities rather than a free-form generator, leaving little room to introduce non-existent ones.
Our manual audit of over 6.8K samples confirms this: only 1.6\% exhibit any hallucination or factual error (\cref{app:curation}), a rate low enough that factual noise is not a meaningful concern for either the training set or the curated test set.

\subsection{Query and Document Image Pool Construction}
In this subsection, we explain how the pools of question and document images were created, including data sources, replacement strategies for unavailable images, and verification procedures to avoid duplication. Images for questions and documents were sourced from separate pools. 
Question images were taken from the M2KR dataset, including OVEN~\cite{hu2023opendomainvisualentityrecognition}, Google Landmark v2~\cite{weyand2020googlelandmarksdatasetv2}, and iNaturalist~\cite{vanhorn2018inaturalistspeciesclassificationdetection}. 
Document images were downloaded from web page links provided by MuKA~\cite{deng-etal-2025-muka}. 
For images that were no longer available, replacements were retrieved using the Wikipedia API and Google Lens, and verified with the imagehash toolkit to ensure they were not identical to the query images. We used a perceptual hash size of 8, treating a distance of 10 or less as identical, following \citet{deng-etal-2025-muka}.

\subsection{Paraphrasing to Prevent Lexical Match}
\label{app:paraphrase}
Here, we describe the paraphrasing process applied to the questions to prevent them from being directly retrievable using the text alone.
We evaluate retrieval using BM25~\cite{robertson1995okapi} with a top-5 setting. Before paraphrasing, approximately 37\% of the noisy samples are retrievable. After paraphrasing, recall drops from 37\% to 25\%, and the remaining retrievable samples are removed from the dataset.

\subsection{Human Verification and Test Set Selection}
\label{app:curation}
We conduct a two-goal human review: (1) auditing the factual quality of pipeline-generated samples, and (2) selecting test samples that meet strict evaluation criteria.
Three fluent English-speaking annotators examined 6.8K randomly sampled questions following a shared written guideline (2.3K per annotator, 24h per split); cross-verification on 150 samples, each independently reviewed by two annotators, yielded 91.3\% agreement.

\noindent\textbf{Quality audit.} \ The audit confirms the reliability of our graph-guided pipeline: 98.4\% of generated samples are factually valid, with only 1.6\% exhibiting hallucination or factual errors.

\begin{figure}[t]
    \centering
    \includegraphics[width=0.95\linewidth]{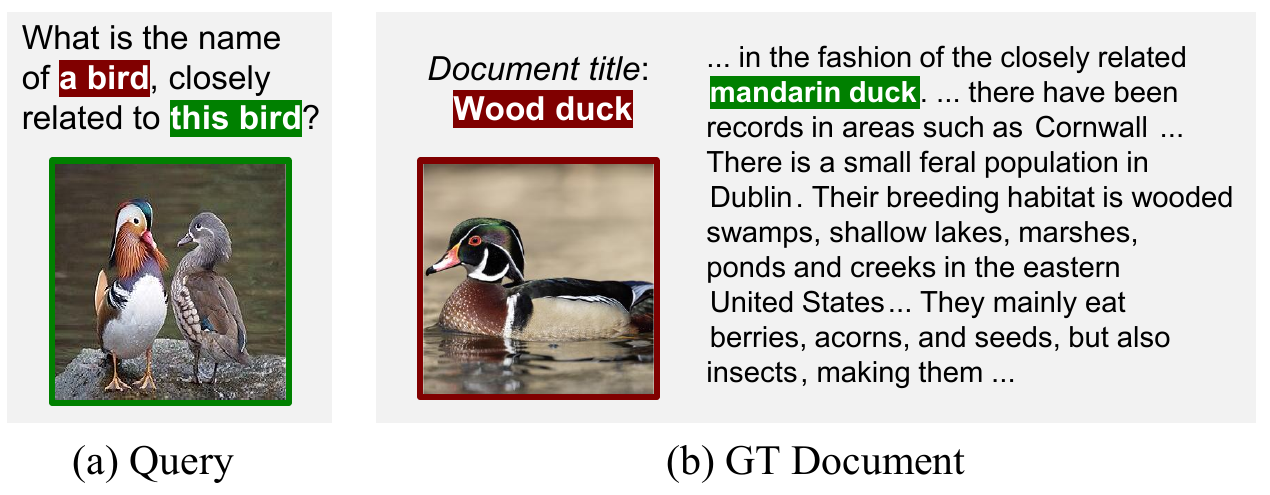}
    \caption{\textbf{Example of a Valid Question Filtered Out due to Evaluation Constraints.} While both the question and its answer are factually valid, multiple duck species may be as closely related to the mandarin duck (green) as the wood duck (red) is. This ambiguity complicates strict automated evaluation, which is why such samples are filtered from the test set.}
    \label{fig:bad_example}
    \vspace{-1em}
\end{figure}

\noindent\textbf{Test set selection.} \ Beyond factual validity, reliable \emph{evaluation} imposes stricter requirements: each test query must admit exactly one correct answer and must not be solvable from a single modality.
Applying these criteria, 29.16\% of audited samples qualified, yielding 1K test samples each for the seen and unseen splits.
We stress that this low pass rate reflects the \emph{stringency of our evaluation criteria}---answer uniqueness and multimodal necessity---not noise in the generation pipeline, whose factual validity is separately established above (98.4\%).
Crucially, samples excluded at this stage are not erroneous---97.74\% of them remain factually valid questions (\Cref{fig:bad_example})---but involve multiple valid answer candidates or potential single-modality solvability, which would confound evaluation.
\added{Annotators selected a rejection reason from a predefined list for every excluded question: multiple correct answers (75.35\%), solvable without the image (21.27\%), lexical cues revealing the answer (2.26\%), non-English text (0.62\%), grammatical errors (0.35\%), and images explicitly revealing the answer (0.15\%). Over 96\% of exclusions thus stem from the two evaluation criteria, answer uniqueness and multimodal necessity, rather than from generation errors; such samples remain correct and are kept as training supervision (\cref{app:train_quality}). Test samples are drawn from a uniformly random audit pool, and train and test questions have similar length statistics (\cref{tab:dataset_stats}).}

\section{Effect of the Qualifying Entity}
\label{app:qualifying}
The qualifying entity is designed to \emph{reduce}, not eliminate, answer ambiguity (\cref{sec:pipeline}): in an automated pipeline, guaranteeing uniqueness is infeasible, which is precisely why the test set undergoes strict human selection (\cref{app:curation}).
Empirically, adding qualifying entities raises the proportion of unambiguous samples from 5\% to 30\%---a six-fold improvement that substantially reduces the human selection burden.
Residual ambiguity in training samples does not harm retrieval supervision (\cref{app:train_quality}), and no ambiguous sample survives into the test set.

\section{Training Set Quality}
\label{app:train_quality}
We emphasize the distinction between \emph{factual validity} and \emph{evaluation strictness}.
Our human audit (\cref{app:curation}) found 98.4\% of pipeline-generated samples to be factually valid; since the audit covered a uniformly random 6.8K subset of the same generation process, this estimate applies to the training set as well.
The strict test criteria---answer uniqueness and multimodal necessity---serve evaluation reliability, not data correctness, and we deliberately do not impose them on training data.
For retrieval training, a question admitting multiple valid answer entities (\eg, ``Which insect feeds on this plant?'') still pairs the query with a genuinely relevant document, providing a correct positive signal; restricting training to strictly unique-answer samples would discard valid supervision and reduce scale and diversity without benefit.
This practice mirrors existing benchmarks: InfoSeek and E-VQA likewise train on automatically generated data at scale while evaluating on cleaner subsets~\cite{chen2023can, mensink2023encyclopedic}.

\begin{table}[t]
\centering
\caption{\textbf{Summary of Dataset Statistics.}}
\label{tab:dataset_stats}
\scalebox{0.75}{ 
\begin{tabular}{lr}
\toprule
\textbf{} & \textbf{Value} \\ 
\midrule
Number of Main Entities & 19,408 \\
Number of Extracted Relations & 66,244 \\
\midrule
Avg. Train Question Length (tokens) & 26.4 \\
Avg. Train Answer Length (tokens) & 5.8 \\
Avg. Test Question Length (tokens) & 28.0 \\
Avg. Test Answer Length (tokens) & 5.5 \\
\bottomrule
\end{tabular}
}
\end{table}

\section{Dataset Statistics}
The statistical properties of our dataset are summarized in Table \ref{tab:dataset_stats}. The graph contains a total of 19,408 unique main entities connected by 66,244 relations. Regarding text length distribution, the training set features an average question length of 26.4 tokens and an average answer length of 5.8 tokens. The test set exhibits a similar distribution, with average question and answer lengths of 28.0 and 5.5 tokens, respectively.

\begin{table*}[t]
\centering
\caption{
\textbf{Standard deviation analysis}. Average retrieval performance and standard deviations of {\modelname} are reported.
}
\scalebox{0.75}{
\begin{tabular}{llcccc}
\toprule
\textbf{Method} & \textbf{Document Modality} 
& \textbf{InfoSeek} 
& \textbf{EVQA} 
& \textbf{{\datasetname} (Seen)} 
& \textbf{{\datasetname} (Unseen)} \\
\specialrule{1pt}{2pt}{2pt}
PreFLMR~\cite{lin2024preflmr} & text & 40.7 & 63.9 & 23.6 & 11.1 \\
UniIR~\cite{wei2024uniir} & text + single-image & 34.5 & 22.0 & 13.8 & 4.7 \\
ReT~\cite{caffagni2025recurrence} & text + single-image & 50.1 & 46.3 & 15.1 & 6.8 \\
MuKA~\cite{deng-etal-2025-muka} & text + single-image & 51.7 & 64.9 & 20.8 & 8.6 \\
\textbf{{\modelname} (Ours)} & \textbf{text + multi-image} & \textbf{53.3} ($\pm 0.5$) & \textbf{65.8} ($\pm 0.4$)& \textbf{43.2} ($\pm 0.5$)& \textbf{36.9} ($\pm 0.6$)\\
\bottomrule
\end{tabular}
}
\label{tab:std}
\end{table*}

\begin{table}[t]
\centering
\caption{
\textbf{{\modelname} with Diverse VLM Backbones.} QA performance (BEM~\cite{bulian-etal-2022-tomayto}) with and without retrieval.
}
\label{tab:vlm_comparison}
\scalebox{0.75}{%
\begin{tabular}{lccc}
\toprule
\textbf{Model} & \textbf{Retrieval} & \textbf{Seen} & \textbf{Unseen} \\
\midrule
\multirow{2}{*}{VILA-13B \cite{lin2024vilapretrainingvisuallanguage}} & \xmark & 11.2 & 10.1 \\
 & \cmark & 32.7 
 & 31.6 
 \\
\midrule
\multirow{2}{*}{InternVL3-8B \cite{zhu2025internvl3}} & \xmark & 18.7 & 17.6 \\
 & \cmark & 41.3 
 & 39.2 
 \\
\midrule
\multirow{2}{*}{Qwen3-VL-8B \cite{yang2025qwen3}} & \xmark & 20.3 & 19.4 \\
 & \cmark & 45.7 
 & 43.9 
 \\
\midrule
\multirow{2}{*}{GPT-4o \cite{hurst2024gpt}} & \xmark & 35.6 & 37.5 \\
 & \cmark & 55.8 
 & 57.2 
 \\
\bottomrule
\end{tabular}
}
\end{table}

\section{Implementation Details}
\label{app:details}
In this section, we provide implementation details, including the dataset construction process and model configurations.
Following MuKA~\cite{deng-etal-2025-muka}, the model is initialized with the weights of PreFLMR~\cite{lin2024preflmr} using the CLIP ViT-G variant, which has demonstrated the best performance. 
Training was conducted for one epoch using the Adam optimizer.
The projection layers are 2-layer MLPs following \citet{deng-etal-2025-muka,lin2024preflmr} for projecting the multimodal features as 32 tokens embeddings and lowering the dimension $\mathbb{R}^{D}$ to $D=128$ for efficiency~\cite{lin2023finegrainedlateinteractionmultimodalretrieval, lin2024preflmr,deng-etal-2025-muka}.
For the number of augmented related images, we do not impose an upper bound on the number of augmented related images, which is an average 4.3 per sample.
For training, we utilize 120k training samples randomly sampled from EVQA and {\datasetname}, respectively.
For the preliminary experiment, we discarded the text token features from the query embeddings while keeping all other architectural components and settings identical~\cite{deng-etal-2025-muka}.
For answer generation, the VILA 13B model~\cite{lin2024vilapretrainingvisuallanguage} employed in \citet{deng-etal-2025-muka} was used. The top-5 results from the zero-shot inference of the PreFLMR~\cite{lin2024preflmr} were utilized to perform LoRA training on VILA 13B. LoRA rank and alpha were set to 128 and 256, respectively, with a batch of 512. 
The Adam optimizer was applied, training only the multimodal projectors and LoRA modules with learning rates of 2e-5 and 2e-4~\cite{deng-etal-2025-muka}.

\section{KB Construction for Inference}

\label{app:kb}
In this section, we describe how the KB is constructed for retrieval, including the passages used and their configurations.  
EVQA and {\datasetname} share the same KB, built from the EVQA subset of Wikipedia passages provided by the M2KR benchmark, while Infoseek uses a separate KB constructed from the Infoseek subset of M2KR Wikipedia passages~\cite{lin2024preflmr}. Since {\datasetname} shares the same KB as EVQA, only samples whose GT documents do not appear in the EVQA training set are included as unseen test samples for {\datasetname}.
For KB indexing with PLAID~\cite{santhanam2022plaidefficientenginelate}, we set the parameters as follows: the number of bits to 8, the number of k-means iterations to 20, used cosine similarity and the batch size was set to 64 following~\citet{santhanam2022colbertv2effectiveefficientretrieval,deng-etal-2025-muka,lin2024preflmr}.

\section{Standard Deviation Analysis}
\label{app:std}
This section analyzes how consistent {\modelname}'s performance is across multiple training runs.
We trained the model 10 times with different random seeds. The resulting standard deviation across datasets was 0.5 on average. 

\section{{\modelname} with Diverse VLM Backbones} \ \ 
To ensure a fair comparison with existing baselines, we primarily adopted VILA-13B as our VLM backbone in previous experiments. 
We further evaluate {\modelname} across a broad spectrum of modern VLMs, including both open-source and proprietary models.
As shown in \cref{tab:vlm_comparison}, while more powerful VLMs generally exhibit higher baseline performance and the performance gap tends to narrow as VLM capability increases, they still struggle significantly to reason over fine-grained entities when relying solely on their internal parameters.
Notably, a substantial performance gap around 20pt persists between with and without {\modelname} even for the strongest VLMs. 
This demonstrates that high-quality retrieved context remains indispensable for bridging the knowledge gaps in even the most advanced VLMs on {\datasetname}.

\begin{table}[t]
    \centering
    \setlength{\tabcolsep}{5pt}
    \scalebox{0.85}{
        \begin{tabular}{lcccc}
        \toprule
        \multicolumn{5}{c}{\textbf{(a) Image selection strategy} (Recall@5)} \\
        \midrule
        \textbf{Strategy} & $K{=}1$ & $K{=}2$ & $K{=}3$ & $K{=}4$ \\
        \midrule
        Random   & 22.5 & 27.8 & 35.9 & 39.2 \\
        Frequent & 24.7 & 28.4 & 38.4 & 41.3 \\
        \midrule
        \multicolumn{5}{c}{\textbf{(b) Noise robustness} (Recall@5)} \\
        \midrule
        \textbf{Noise rate} & 0\% & 10\% & 20\% & 30\% \\
        \midrule
        {\modelname} & 43.2 & 42.8 & 40.9 & 36.8 \\
        \bottomrule
        \end{tabular}
    }
    \caption{\textbf{Image Selection and Noise Robustness} on {\datasetname}. (a) Both selection strategies improve consistently with $K$, with frequency-based selection slightly ahead. (b) {\modelname} degrades gracefully under noisy related-entity images, remaining stable up to 20\%. \added{Noise replaces the given fraction of related-entity images with images of unrelated entities while keeping the number of images per document, the architecture, and training fixed, so it varies only the correctness of the entity association.}}
    \vspace{-1em}
    \label{tab:selection_noise}
\end{table}

\noindent\textbf{Image Selection and Noise Robustness} \ \
While {\modelname} augments documents with images of all related entities, we simulate settings where related images are missing or noisy (\cref{tab:selection_noise}).
We compare two selection strategies: randomly selecting $K$ related entities, and prioritizing the $K$ most frequent entities in the KB.
Even at $K{=}1$, both strategies outperform MuKA, and performance scales with $K$, confirming that broader entity coverage benefits.
Under noise injection at 10\%, 20\%, and 30\%, {\modelname} remains stable up to 20\%, demonstrating resilience to imperfect entity images.
\added{Because the number of indexed images is fixed under noise injection, the monotonic decline also serves as a capacity-controlled test: performance depends on whether the indexed images depict genuinely related entities, not on how many are indexed (\cref{sec:capacity}).}

\section{Efficiency Analysis and Model Complexity} 
\label{app:efficiency}
At inference, document representations are pre-indexed; only the query is encoded on-the-fly for MaxSim computation (\cref{sec:late_interaction}).
Although the additional images and multimodal tokens increase the tokens in MaxSim by $2.3\times$, the impact on end-to-end latency is negligible: query encoding (\cref{sec:query_emb}) dominates the runtime, being over $1000\times$ slower than MaxSim.
\added{This holds because PLAID~\cite{santhanam2022plaidefficientenginelate} prunes candidates through centroid interaction and computes full late-interaction scores only for a small candidate set, so the additional document tokens mainly enlarge the index rather than the per-query scoring cost. Measured per query, encoding takes $\sim$2\,s while retrieval takes $\sim$0.002\,s, and the query encoder is identical for {\modelname} and the single-image baseline.}
During training, the vision encoder remains frozen, allowing for the pre-extraction of features and resulting in training times of 2.5 hrs for the single-image baseline and 4.4 hrs for MIMIR on a single A100 GPU.

\section{\added{Comparison with Multimodal KG-RAG}}
\label{app:kgrag}
\added{We compare {\modelname} with two multimodal knowledge-graph RAG methods, RAG-Anything~\cite{guo2025raganything} and mKG-RAG~\cite{yuan2025mkgrag}, on the same KB as \cref{tab:main}.
Both methods retrieve entities and relations from a graph rather than documents, so we map the retrieved entities and relations back to their source documents and rank documents by their graph retrieval scores to compute R@5 (\cref{tab:kgrag}). Note that mKG-RAG originally evaluates InfoSeek on an EchoSight-filtered subset; we instead use the M2KR InfoSeek KB for consistency with all other methods, so the numbers are not directly comparable to those reported in the mKG-RAG paper.}
\added{{\modelname} outperforms both methods on every dataset and split, improving R@5 averaged over the four evaluation sets by 94\% and 53\% relative to RAG-Anything and mKG-RAG, respectively. Structured retrieval does not by itself resolve the visual shortcut: mKG-RAG builds a query-specific graph only from documents recalled by a single-image document retriever, so when the query image depicts a related entity, the GT document is often never recalled and never enters graph construction.
We further replace this document-recall stage with MuKA or {\modelname}. Both improve over standalone mKG-RAG (by 6\% and 14\% in relative average R@5, respectively), yet mKG-RAG + {\modelname} remains 25\% below standalone {\modelname}: since {\modelname} already captures most relevant candidates, the graph stage mostly fetches redundant nodes or introduces noise from incomplete graph extraction.}

\begin{table}[t]
\centering
\revcolor
\setlength{\tabcolsep}{4pt}
\resizebox{\linewidth}{!}{
\begin{tabular}{lcccc}
\toprule
& \textbf{InfoSeek} & \textbf{EVQA} & \multicolumn{2}{c}{\textbf{{\datasetname}}} \\
\cmidrule(lr){4-5}
\textbf{Method} & & & Seen & Unseen \\
\midrule
RAG-Anything~\cite{guo2025raganything} & 34.0 & 27.7 & 21.3 & 19.2 \\
mKG-RAG~\cite{yuan2025mkgrag} & 45.7 & 35.8 & 25.8 & 22.8 \\
mKG-RAG + MuKA & 48.4 & 37.2 & 28.2 & 24.9 \\
mKG-RAG + {\modelname} & 47.9 & 38.6 & 32.6 & 29.3 \\
\midrule
\textbf{{\modelname}} & \textbf{53.3} & \textbf{65.8} & \textbf{43.2} & \textbf{36.9} \\
\bottomrule
\end{tabular}
}
\caption{\added{\textbf{Comparison with Multimodal KG-RAG Methods} (R@5) on the same KB under a shared top-5 retrieval protocol.}}
\label{tab:kgrag}
\end{table}

\section{Dataset Release Details}
\label{app:release}
\noindent\textbf{Released artifacts.} \ We will publicly release the following:
\begin{itemize}[noitemsep, topsep=2pt, leftmargin=*]
    \item \textbf{{\datasetname} training set} (120K samples) and \textbf{human-curated test set} (2K samples, with seen/unseen splits). Each sample contains: the query image identifier, the textual question, the answer, the GT document identifier, and the intermediate pipeline annotations (main entity, query entity, qualifying entity, and the relation sentences of the target subgraph).
    \item \textbf{Knowledge base augmentation files}: the mapping from each document to its related entities and their image identifiers, enabling reconstruction of the multi-image augmented KB described in \cref{sec:doc_emb}.
    \item \textbf{Construction pipeline}: code and prompts for the graph construction, subgraph selection, and QA generation stages, enabling reproduction and extension to other KBs.
    \item \added{\textbf{{\modelname} code and checkpoints}: training and inference code, PLAID indexing scripts, and trained model weights.}
\end{itemize}

\noindent\textbf{Format.} \ Samples are distributed as JSON Lines and Parquet files with a documented schema; images are referenced by M2KR identifiers and Wikimedia URLs rather than redistributed, respecting the licensing of the source images.

\noindent\textbf{Source data and licensing.} \ {\datasetname} is built from Wikipedia text (CC BY-SA 4.0) and images from M2KR~\cite{lin2024preflmr}, inheriting their license terms. Our annotations and generated questions (produced with Qwen-2.5-32B-Instruct~\cite{bai2023qwen}) are released under CC BY 4.0. The dataset is intended for research use in multimodal retrieval and KB-VQA.

\noindent\textbf{Hosting.} \ The dataset will be hosted at Hugging Face Datasets with versioned releases and documentation of the curation criteria (\cref{app:curation}).

\section{Robustness to Visual-similar Distractors}
\label{app:distractor}
\begin{figure}[t]
    \centering
        \begin{subfigure}{0.49\linewidth}
        \centering
        \includegraphics[width=0.9\linewidth]{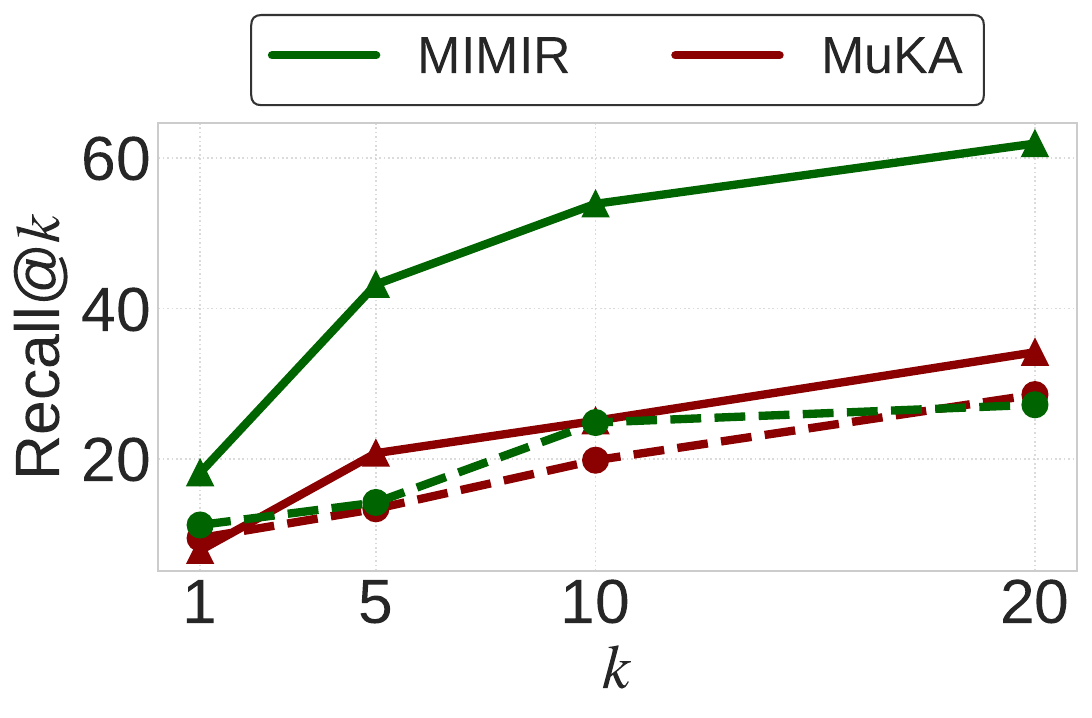}
        \caption{Seen}
        \label{fig:distractor-seen}
    \end{subfigure}
    \hfill
    \begin{subfigure}{0.49\linewidth}
        \centering
        \includegraphics[width=0.9\linewidth]{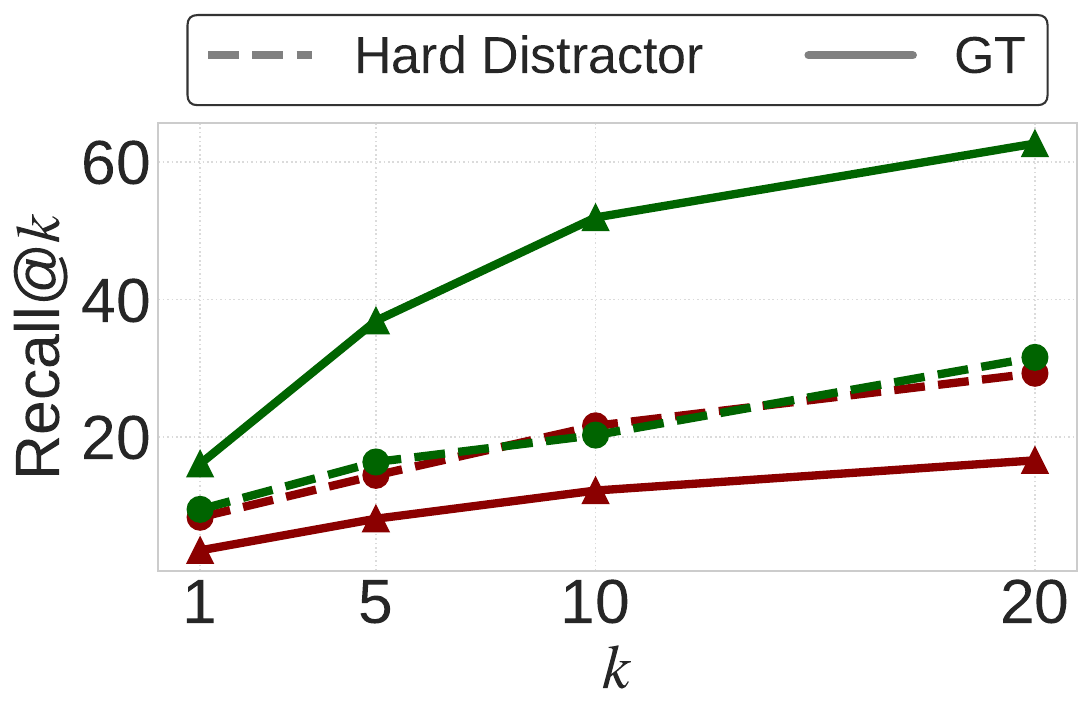}
        \caption{Unseen}

        \label{fig:distractor-unseen}
    \end{subfigure}
    \caption{
    \textbf{Robustness to Visual-similar Distractors.} {\modelname} (\textbf{\textcolor{figgreen}{green}}) and MuKA (\textbf{\textcolor{figbrightred}{red}}) curves show recall for distractor (dotted) and GT (solid) documents. Distractors are non-GT documents whose main entity matches the query image’s entity. 
    }
    \label{fig:distractor_recall}
\end{figure}

We analyze {\modelname} and MuKA, both trained and tested on {\datasetname}, against hard distractors: non-GT documents whose main entity matches the query image but lacks the queried information.
\Cref{fig:distractor_recall} reports Recall@$k$ for GT documents (solid) and distractors (dotted).
For MuKA, the two curves stay close---at Recall@1, distractors are retrieved more often than GT documents, worsening on the unseen subset---and in 11.3\% of all cases (78\% of distractor retrievals), the GT document falls outside the top-5, reflecting the visual-shortcut bias of single-image augmentation.
Since these distractors match the query image more than the GT document, {\modelname}'s preference for the GT document shows it also relies on text, not visual similarity alone.

\begin{figure*}[t]
  \centering
    \includegraphics[width=\linewidth]{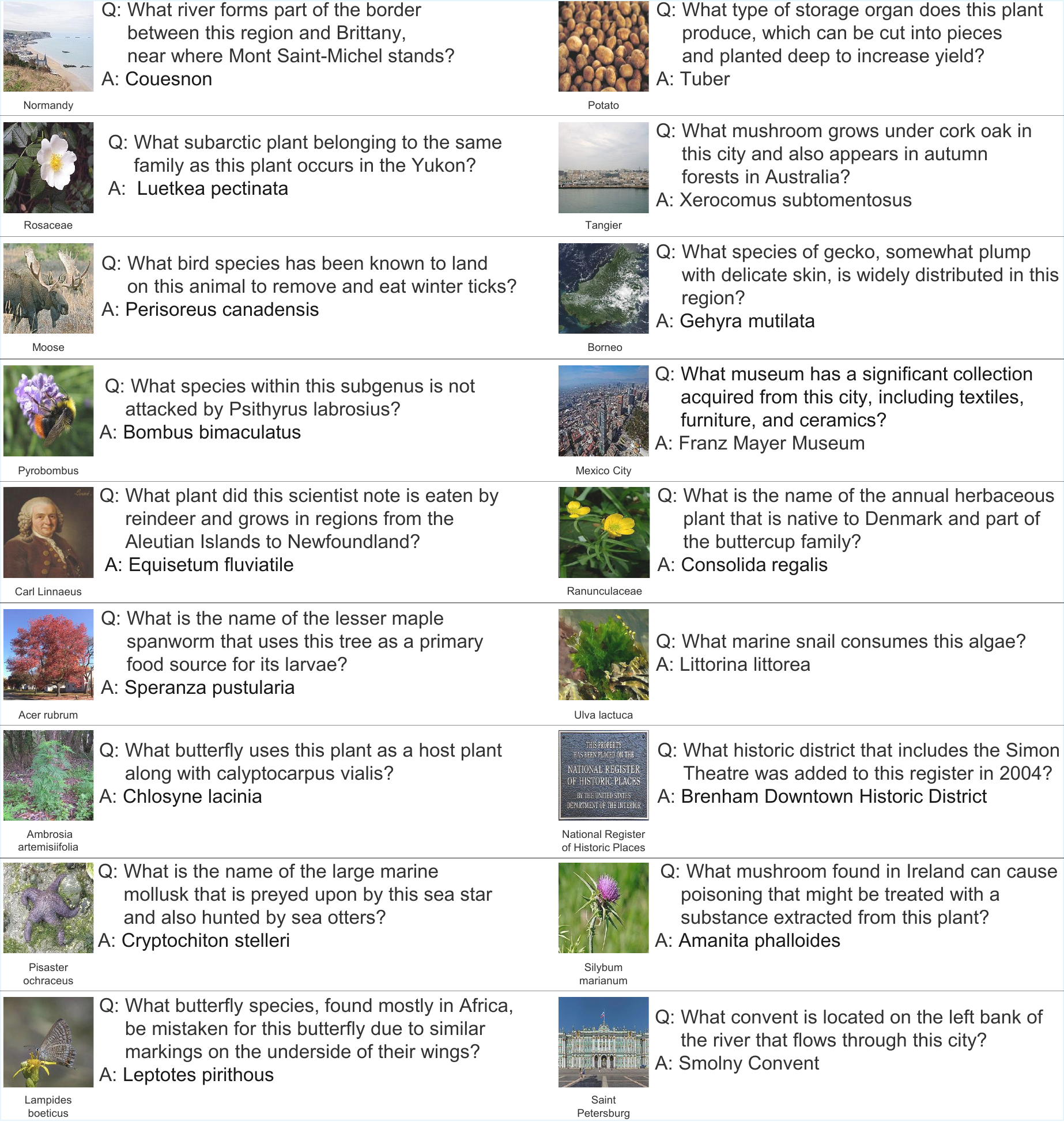}
  \caption{\textbf{Qualitative Examples of {\datasetname} Training Set before Human Curation.}}
  \label{fig:qualitative_qa}
\end{figure*}

\section{Qualitative Examples of {\datasetname}}
\label{app:qualitative_qa}
\Cref{fig:qualitative_qa} presents representative samples generated by our pipeline.
Each sample pairs a query image depicting a related entity with a question that references it via a demonstrative phrase (\eg, ``this reservoir'') and is answered by the main entity of the GT document.
The examples illustrate the diversity of the generated data across entity types and relation structures.
As the query images never depict the GT document's main entity, none of these samples can be resolved through the visual shortcut.

\section{Use of LLMs}
We acknowledge that LLMs were used as writing
assistants to improve grammar, clarity, and readability of the manuscript.

\begin{figure*}[t]
\begin{tcolorbox}[colframe=gray!50!black, colback=gray!10!white, title=Entity extraction prompt, width=\textwidth, sharp corners=southwest]
\scriptsize
\begin{verbatim}
You are a Named Entity Recognition and Relation Extraction system specialized for 
identifying entities that
correspond to Wikipedia article title.

Your task is to extract all distinct named entities mentioned or referenced in the 
passage that align with
Wikipedia article titles, even if they are not exact substrings.

Guidelines:
1. For each entity, return an object with the fields: "entity", "entity_type", 
and "relation".
2. The "entity" must be a canonical Wikipedia title or a close variant that accurately 
represents the mention 
in the passage.
3. "entity_type" describes the category of the entity (e.g., Person, Place, 
Organization, Event, Book, Law, etc.).
You may create new types as needed.
4. The "relation" must follow these strict rules:
   - It must be a single, full natural language sentence.
   - The sentence must explicitly include the main entity as the subject.
   - It must describe concrete and fact-based relationships to the given entity, 
   including specific information 
   such as dates, functions, locations, roles, historical context, or other verifiable 
   details.
   - If there are multiple distinct factual relations for the entity, combine them 
   into one coherent 
   and grammatically correct sentence.
   - The sentence must be informative enough to support downstream question generation 
   tasks.
   - Avoid abstract, vague, or generic relations (e.g., "is related to", "is part of", 
   or incomplete phrases).
5. If the passage does not provide any clear relation for the entity, set "relation" 
to null.
6. Merge variations or aliases of the same entity to avoid duplication.
7. Be comprehensive and tolerant of indirect mentions or paraphrased references.

Wikipedia Title: "{entity}"

Passage:
\end{verbatim}
\end{tcolorbox}
\caption{\textbf{Entity Extraction Prompt.}}
\label{fig:entity_extraction_prompt}
\end{figure*}
\begin{figure*}[t]
\begin{tcolorbox}[colframe=gray!50!black, colback=gray!10!white, title=Question Generation Prompt, width=\textwidth, sharp corners=southwest]
\scriptsize
\begin{verbatim}
You are a VQA question-answer pair generator.

Given:
- Answer entity (answer to the question)
- Query entity (depicted in the image)
- Query entity's relation sentence describing how main and related entities 
are connected
- Qualifying entity's relation sentence describing the main entity

Generate a single, grammatically correct question-answer pair satisfying:

- The question refers indirectly to the query entity shown in the image by including 
exactly one singular 
demonstrative phrase matching the query entity’s type, e.g. "this country", 
"this building", "this city", etc.
- Do NOT explicitly name the query entity in the question.
- The demonstrative phrase must appear exactly once and without any article 
(no "a", "an", or "the").
- The question must use only the query entity's relation and/or qualifying entity's 
relation to describe the answer entity indirectly.
- The answer must be exactly the answer entity’s name.
- Do NOT mention or describe the image explicitly 
(no "shown here", "in this image", etc.).
- Output the question and answer in the format:

Question: [your question]  
Answer: [main entity]
\end{verbatim}
\end{tcolorbox}
\caption{\textbf{Question Generation Prompt.}}
\label{fig:qa_gen}
\end{figure*}
\begin{figure*}[t]
\begin{tcolorbox}[colframe=gray!50!black, colback=gray!10!white, title=Paraphrasing Prompt, width=\textwidth, sharp corners=southwest]
\scriptsize
\begin{verbatim}
You are a helpful assistant that rewrites trivia questions only.  
Keep the original meaning and difficulty but change the wording.  
Do NOT include the answer in your response — output ONLY the rewritten question.  

Rewrite the following location-based trivia question using different wording,  
but keep the original meaning and difficulty level.  

Output format:  
Question: [your rewritten question]  
\end{verbatim}
\end{tcolorbox}
\caption{\textbf{Paraphrasing Prompt.}}
\label{fig:paraphrase}
\end{figure*}
\begin{figure}[t]
\centering
\begin{tcolorbox}[colframe=gray!50!black, colback=gray!10!white,
  title=VILA Training Prompt, width=\columnwidth, sharp corners=southwest,
  boxsep=2pt, left=2pt, right=2pt, top=2pt, bottom=2pt]
\scriptsize
\begin{alltt}
<image>
Question: {question}

Retrieved passages:
1: <image>{document_text_1}
2: <image>{document_text_2}
3: <image>{document_text_3}
4: <image>{document_text_4}
5: <image>{document_text_5}

Given the query image and question,
identify matched passages and provide a short answer.
\end{alltt}
\end{tcolorbox}
\caption{\textbf{VILA Training Prompt.}}
\label{fig:vila_prompt}
\vspace{-0.5em}
\end{figure}

\begin{figure}[t]
\centering
\begin{tcolorbox}[colframe=gray!50!black, colback=gray!10!white,
  title=Retriever Instruction Prompts, width=\columnwidth, sharp corners=southwest,
  boxsep=2pt, left=2pt, right=2pt, top=2pt, bottom=2pt]
\scriptsize
\begin{alltt}
 <Image> Using the provided image, obtain
documents that address the subsequent question:

<Image> Retrieve documents that provide an
answer to the question alongside the image: 

<Image> Extract documents linked to the
question provided in conjunction with the image:

<Image> Utilizing the given image, obtain
documents that respond to the following question:

<Image> Using the given image, access documents that 
provide insights into the following question: 

<Image> Obtain documents that correspond to
the inquiry alongside the provided image: 

<Image> With the provided image, gather documents 
that offer a solution to the question: 

<Image> Utilizing the given image, obtain
documents that respond to the following question:
\end{alltt}
\end{tcolorbox}
\caption{\textbf{Retriever Instruction Prompts.}}
\label{fig:preflmr_prompt}
\vspace{-1.5em}
\end{figure}
\begin{figure*}[t]
  \centering
    \includegraphics[width=0.8\linewidth]{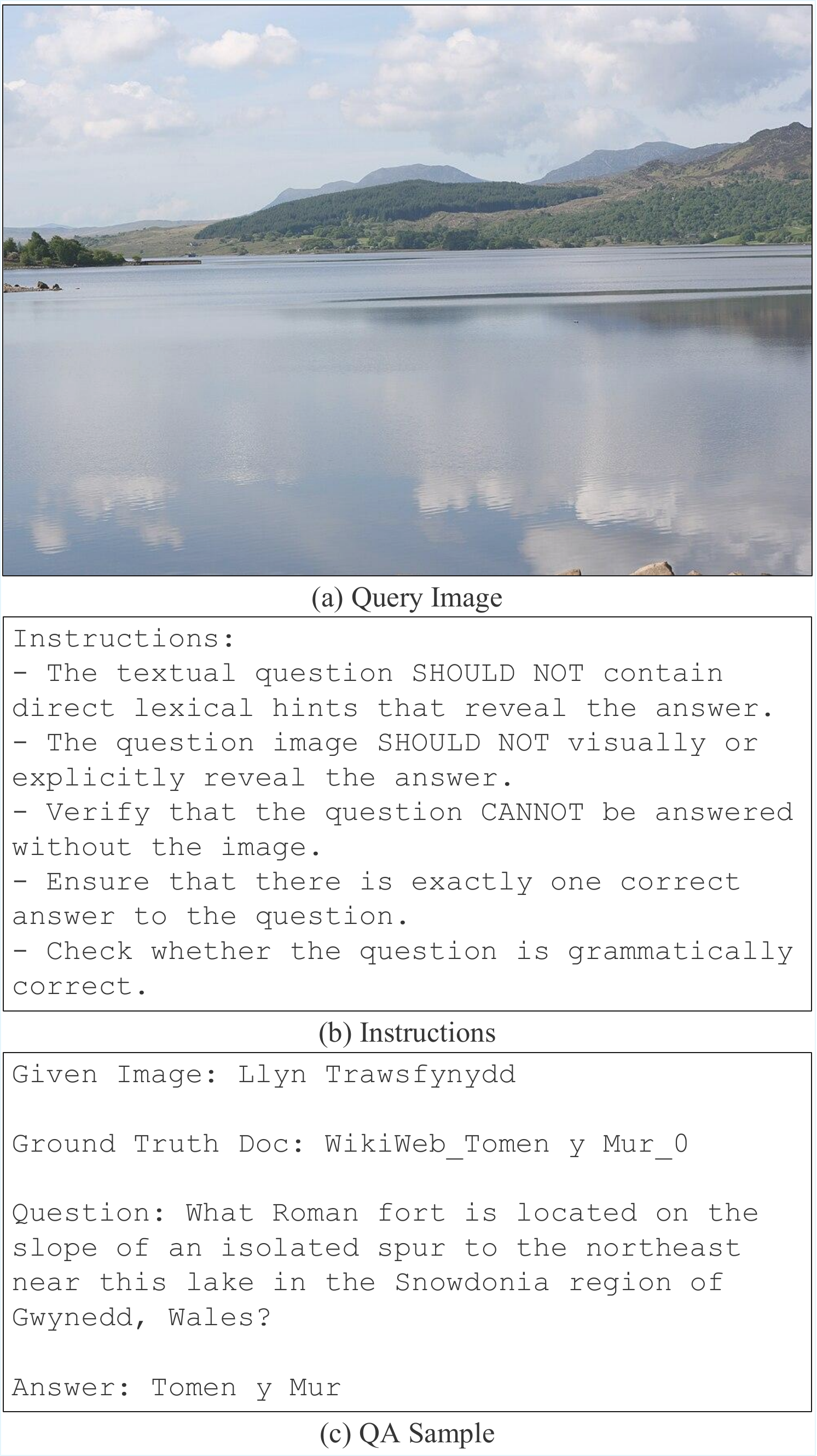}
  \caption{
  \textbf{Human Curation User Interface} 
  An image, document title, a question, and an answer are presented for curation, along with guidelines for identifying common issues.
  }
  \label{fig:ui}
\end{figure*}

\clearpage

\end{document}